\documentclass[10pt,twocolumn,letterpaper]{article}

\usepackage{subcaption}
\usepackage{iccv}
\usepackage{times}
\usepackage{epsfig}
\usepackage{graphicx}
\usepackage{amsmath}
\usepackage{amssymb}
\usepackage{booktabs}
\usepackage{diagbox}
\usepackage[pagebackref=true,breaklinks=true,colorlinks,bookmarks=false]{hyperref}
\usepackage[accsupp]{axessibility}

\usepackage[capitalize]{cleveref}

\iccvfinalcopy 

\begin{document}

\title{Score-Based Diffusion Models as Principled Priors for Inverse Imaging}
\author{Berthy T. Feng$^{1*}$ \quad Jamie Smith$^{2}$ \quad Michael Rubinstein$^{2}$ \quad Huiwen Chang$^{2}$ \\
Katherine L. Bouman$^{1}$  \quad William T. Freeman$^{2}$\\
{$^{1}$California Institute of Technology \quad $^{2}$Google Research}
}

\twocolumn[{%
\renewcommand\twocolumn[1][]{#1}%
\maketitle
\begin{center}
  \centering
  \captionsetup{type=figure}
  \includegraphics[width=0.99\textwidth]{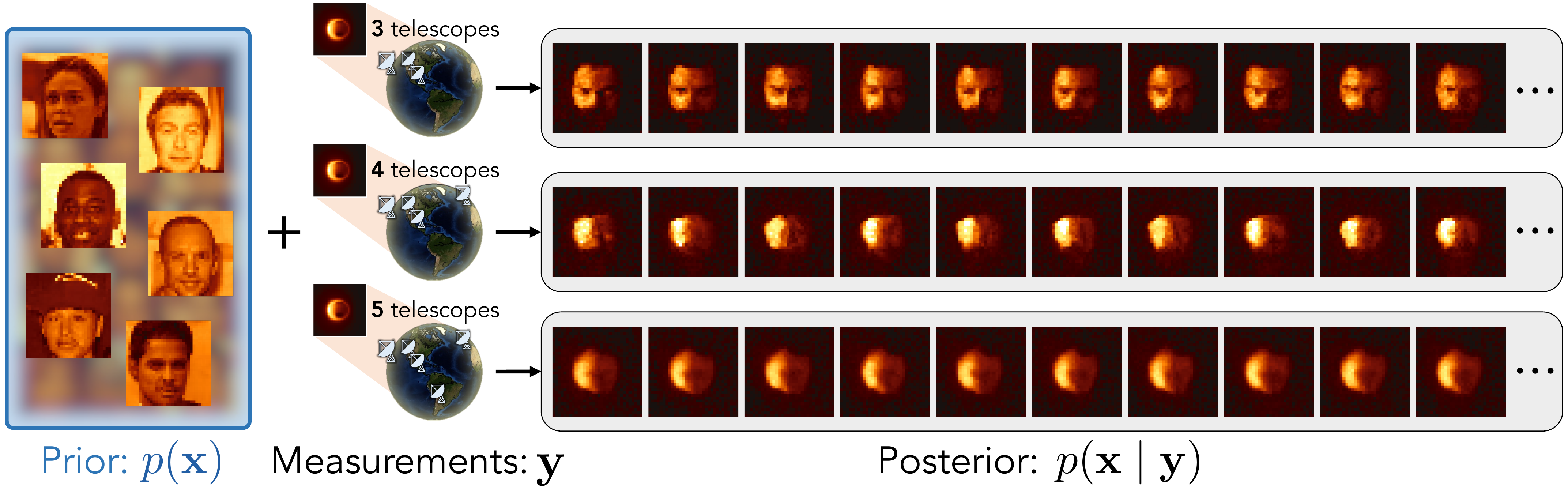}
  \captionof{figure}{
  A score-based prior is a \textbf{hyperparameter-free, probabilistic prior} that is also \textbf{expressive and data-driven}. Paired with a set of measurements, the prior can be used for principled inference of a full posterior. In this example, a score-based prior was trained on face images (``Prior'' shows samples from the learned prior). The inverse problem is interferometic imaging of a synthetic black hole. We simulated interferometric measurements from the actual telescope array used to capture the first black-hole image~\cite{event2019first} and sampled images from the posterior via variational inference.
  From the top to bottom row, the posterior stably moves away from the prior given more constraining measurements. 
  With measurements from only three telescopes, the posterior shows strong influence from the prior and contains images resembling faces that are brighter on the left half. As more telescopes (measurements) are added, the posterior reveals the ring-like structure of the underlying image. Our framework finds the proper relative strengths of the prior and measurements automatically.\vspace{0.1in}
  }
  \label{fig:teaser}
\end{center}%
}]

\ificcvfinal\thispagestyle{empty}\fi

\begin{abstract}
{\let\thefootnote\relax\footnote{{*Work partially done during an internship at Google Research.}}}
Priors are essential for reconstructing images from noisy and/or incomplete measurements. The choice of the prior determines both the quality and uncertainty of recovered images.
We propose turning score-based diffusion models into principled image priors (``score-based priors'') for analyzing a posterior of images given measurements. Previously, probabilistic priors were limited to handcrafted regularizers and simple distributions.
In this work, we empirically validate the theoretically-proven probability function of a score-based diffusion model.
We show how to sample from resulting posteriors by using this probability function for variational inference. Our results, including experiments on denoising, deblurring, and interferometric imaging, suggest that score-based priors enable principled inference with a sophisticated, data-driven image prior.
\end{abstract}

\section{Introduction}
\label{sec:intro}
\vspace{0.1in}
Priors are crucial for solving inverse problems in computational imaging, which tend to be ill-posed due to noisy and limited sensors.
When many different images agree with observed measurements, a prior helps constrain solutions according to desired image statistics. How to incorporate a sophisticated prior, however, is not straightforward. Our work addresses the problem of incorporating a rich prior into principled approaches to inverse problems.

Previous work poses a tradeoff: using principled methods requires simple priors, while using deep-learned priors precludes precise analysis. On the principled side, Bayesian-inference methods model the posterior distribution of images $\mathbf{x}$ conditioned on measurements $\mathbf{y}$:
\begin{align*}
    p(\mathbf{x}\mid\mathbf{y})\propto p(\mathbf{y}\mid\mathbf{x})\:p(\mathbf{x}).
\end{align*}
This Bayesian framework supports a modular approach to inverse problems where the likelihood $p(\mathbf{y}\mid\mathbf{x})$ is defined by an expert based on knowledge of how measurements are obtained, and the prior $p(\mathbf{x})$ is defined independently. Furthermore, it allows for principled solutions. Maximum a posteriori (MAP) estimation can be done by optimizing the posterior probability. Posterior sampling, which is useful for uncertainty quantification, can be done with MCMC or variational inference. But since such methods require the value or gradient of $p(\mathbf{x})$, they have been limited to simple priors (e.g., Gaussian) and weighted regularizers (e.g., total variation).
In practice, the relative weights of the prior and likelihood terms are usually tuned by hand, introducing a human bias that is unsatisfactory for scientific applications.

On the deep-learning side, solutions leveraging an implicit, deep-learned prior may look convincing but do not lend themselves to principled analysis. For example, a convolutional neural network (CNN) can be trained in a supervised way to output images given measurements, but its prior cannot be probed and does not generalize to new tasks. Recent work shows how to condition a diffusion model --- a type of generative model whose prior is captured in a learned image denoiser --- on arbitrary measurements~\cite{chung2023diffusion,chung2022improving,chung2022come,chung2022score,graikos2022diffusion,song2023pseudoinverseguided,song2022solving}, but the methods depend on hand-tuned hyperparameters and do not sample from a true posterior except in auspicious cases.
To get the best of both worlds (traditional Bayesian inference and modern deep learning), we need a way to incorporate the expressive prior of a deep-learned model into a traditional, principled Bayesian-inference approach.

We propose employing a diffusion model as the prior in Bayesian inference for imaging. A \textit{score-based prior} is the distribution under a score-based diffusion model~\cite{song2021scorebased}, which has been proven to allow for exact probabilities (a feature under-explored in practice). In this paper, we first review related work and then investigate the probability function, including empirical validation of its accuracy. 

The main contribution of this paper is establishing score-based priors as an interface between modern deep-learning and traditional inverse problem-solving, giving proven, principled approaches direct access to learned, rich priors. Under our framework, we train a score-based prior once on a dataset of images.
Paired with any likelihood $p(\mathbf{y}\mid\mathbf{x})$, this prior can be plugged into any inference algorithm that uses the value or gradient of the posterior.
We demonstrate this with an existing variational-inference approach for posterior sampling and show results for three inverse problems: denoising, a version of deblurring, and interferometry. Interferometry is used for black-hole imaging (Fig.~\ref{fig:teaser}) and highlights the benefits of score-based priors for scientific applications, which call for exact posterior sampling given standalone priors to accurately quantify uncertainty.

\section{Related Work}
\label{sec:related_work}
\subsection{Inverse Problems in Imaging}
The goal of an imaging inverse problem is to recover a hidden image $\mathbf{x}^*\in\mathbb{R}^D$ from measurements $\mathbf{y}\in\mathbb{R}^M$:
\begin{align}
\label{eq:forward_model}
    \mathbf{y}=\mathbf{f}(\mathbf{x}^*)+\mathbf{\epsilon}.
\end{align}
We usually assume the forward model $\mathbf{f}$ is known and the measurement noise $\mathbf{\epsilon}$ is a random variable with a known distribution. When $\mathbf{y}$ is missing information about the underlying image, solving for $\mathbf{x}^*$ is ill-posed.

\textbf{Bayesian inference.} The Bayesian approach considers the posterior distribution $p(\mathbf{x}\mid\mathbf{y})$, which decomposes an inverse problem explicitly into a likelihood and a prior:
\begin{align}
\label{eq:logposterior}
    \log p(\mathbf{x}\mid\mathbf{y})=\log p(\mathbf{y}\mid\mathbf{x})+\log p(\mathbf{x})+ \text{const.}
\end{align}
This quantity is often re-interpreted as a data-fidelity term plus a weighted regularizer. Possible solutions to an inverse problem include the maximum a posteriori (MAP), which is the mode of the posterior, and unbiased samples from the posterior. Samples are more informative than the MAP and allow for uncertainty quantification. Principled approaches for posterior sampling include Markov chain Monte Carlo (MCMC)~\cite{brooks2011handbook}, which generates a Markov chain whose stationary distribution is the posterior, and variational inference~\cite{blei2017variational}, which finds the best approximation of the posterior within a family of parameterized distributions. Such algorithms require either the value or gradient of the posterior density function, which is especially difficult to determine for images. Assuming a known likelihood, the challenge is defining a prior on images that reflects their complicated statistics. As such, image priors are usually regularizers enforcing some simple property of images. Examples include total variation (TV) and total squared variation (TSV) for spatial smoothness~\cite{bouman1993generalized,kuramochi2018superresolution} and L1 norm for sparsity~\cite{candes2007sparsity}.  The weightings of these regularizers are typically set by hand.

\textbf{Deep learning for inverse problems.} Deep neural networks can learn complex image distributions, but their implicit priors are difficult to analyze.
One could train a neural network on a paired dataset of images and measurements~\cite{pathak2016context,iizuka2017globally,yu2019free,zhang2022deep,zhang2020extending,yin2021end,saharia2022image,delbracio2021projected,delbracio2023inversion}, but this requires re-training for new tasks, and uncertainty cannot be analyzed under the learned prior. Bayesian networks~\cite{gal2015bayesian} can account for uncertainty, but their priors are still implicit, and they have not been demonstrated on complicated posteriors. Deep Image Prior~\cite{ulyanov2018deep} showed that the inductive bias of a CNN can act as an implicit prior. Other methods like Plug-and-Play (PnP)~\cite{venkatakrishnan2013plug} and Regularization by Denoising (RED)~\cite{romano2017little} use an image denoiser as an implicit regularizer to provide a point solution. PnP/RED-style methods that provide samples~\cite{laumont2022bayesian,kadkhodaie2021stochastic,kawar2021snips} have not been shown to sample from true posteriors based on the denoiser's prior.

\subsection{Data-Driven Generative Models}
Generative models learn a probability distribution of images, but besides score-based diffusion models, they have been either limited in complexity or unable to provide exact image probabilities. Classical examples are Gaussian mixture models~\cite{zoran2011learning,zoran2012natural} and independent components analysis (ICA)~\cite{bell1997independent,hyvarinen2000independent}, which give probabilities but over-simplify image distributions. As for deep generative models, generative adversarial networks (GANs)~\cite{goodfellow2020generative} do not have tractable probabilities, and variational autoencoders (VAEs)~\cite{kingma2013auto,rezende2014stochastic} only provide a lower-bound on probabilities. Flow-based models like discrete normalizing flows~\cite{kingma2018glow,dinh2016density,ho2019flow++,grathwohl2018ffjord,behrmann2019invertible,dinh2014nice,nalisnick2018deep} and autoregressive flows~\cite{larochelle2011neural,germain2015made,van2016pixel,papamakarios2017masked} support exact log-probability computation~\cite{asim2020invertible} but are restricted to certain network architectures and do not generalize well outside of training data~\cite{kirichenko2020normalizing,nalisnick2018deep}.
Diffusion models~\cite{ho2020denoising,kingma2021variational,weng2021diffusion,song2020denoising,nichol2021improved} are a promising alternative, with the score-based interpretation~\cite{song2021scorebased} providing a theoretical framework for deriving exact probabilities~\cite{lu2022maximum}. This capability has gone under-examined in previous work, which mostly focuses on diffusion models as unconditional samplers~\cite{dhariwal2021diffusion,ho2022cascaded,rombach2022high} and conditional samplers~\cite{saharia2022photorealistic,nichol2021glide,ramesh2022hierarchical,ho2022classifier,kawar2022enhancing}.

\subsubsection{Diffusion models}
A diffusion model transforms a simple distribution into a complex one, with image generation usually done by gradually denoising a sample from $\mathcal{N}(\mathbf{0}_D,\mathbf{I}_D)$ until it becomes a clean image. Denoising diffusion probabilistic models (DDPMs)~\cite{sohl2015deep,ho2020denoising} treat this transformation as a discrete-time process with a fixed number of denoising steps. Score-based generative models~\cite{song2019generative} generalize to continuous time.

\textbf{Score-based diffusion models.} In the continuous-time setting, a stochastic differential equation (SDE) describes the data-transformation process and lends itself to a theoretical framework for analyzing the induced sequence of data distributions. In particular, a \textit{forward-time SDE} defines the diffusion process of an image $\mathbf{x}_0$ from $t=0$ to $t=T$: 
\begin{align}
\label{eq:forward}
    \mathrm{d}\mathbf{x}_t=\mathbf{f}(\mathbf{x}_t,t)\mathrm{d}t+g(t)\mathrm{d}\mathbf{w}.
\end{align}
$\mathbf{f}(\cdot, t):\mathbb{R}^D\to\mathbb{R}^D$ is the drift coefficient and defines the deterministic evolution of $\mathbf{x}_t$. $\mathbf{w}\in\mathbb{R}^D$ denotes Brownian motion, and the diffusion coefficient $g(\cdot):\mathbb{R}\to\mathbb{R}$ controls the rate of diffusion. This SDE gives rise to a time-dependent probability distribution $p_t$, where $p_0$ is the original data distribution, and $p_T$ is the standard normal.

Sampling an image from $p_0$ requires reversing the diffusion process. The \textit{reverse-time SDE}~\cite{ANDERSON1982313} is given by
\begin{align}
\label{eq:reverse}
   \mathrm{d}\mathbf{x}_t=\left[\mathbf{f}(\mathbf{x}_t,t)-g(t)^2\nabla_{\mathbf{x}_t}\log p_t(\mathbf{x}_t)\right]\mathrm{d}t + g(t)\mathrm{d}\bar{\mathbf{w}}. 
\end{align}
The problematic term here is $\nabla_{\mathbf{x}}\log p_t(\mathbf{x})$, which is the (Stein) score of $\mathbf{x}$ under $p_t$. In words, undoing diffusion is difficult because it requires knowing how to nudge each perturbed distribution $p_t$ closer to the clean data distribution. Being the only data-dependent component of this SDE, the score function is learned by a neural network $\mathbf{s}_\theta(\mathbf{x},t)$ with parameters $\theta$~\cite{song2021scorebased, song2021maximum}.
Because each $p_t$ is $p_0$ perturbed by Gaussian noise, the time-dependent score model $\mathbf{s}_\theta(\mathbf{x},t)$ can be thought-of as an image denoiser: it takes a noisy image as input and estimates the noise, and $t$ indicates the level of noise applied (higher $t$ means more noise).
To sample a new image, a point $\mathbf{x}_T\sim\mathcal{N}(\mathbf{0}_D,\mathbf{I}_D)$ is drawn
whose final state $\mathbf{x}_0$ is given by solving the reverse-time SDE, essentially through many denoising steps~\cite{song2021scorebased}.

\textbf{Diffusion models for inverse problems.}
Previous methods attempt to sample from a posterior using an unconditional diffusion model but are not guaranteed to sample from the true posterior or a close approximation of it. Most methods incorporate measurements into the reverse-diffusion process. Measurement-consistency is enforced by either projecting images onto a measurement subspace~\cite{song2022solving,chung2022come,chung2022score,choi2021ilvr,chung2022improving} or following a gradient toward higher measurement likelihood~\cite{chung2023diffusion,jalal2021robust,graikos2022diffusion,kawar2022denoising,adam2022posterior}. These methods (1) require careful hyperparameter-tuning to obtain reasonable samples and (2) fail to sample from the true posterior no matter the hyperparameter values. We put forth a new perspective: applying the hyerparameter-free probability function of a score-based diffusion model to principled inference algorithms that simply require a differentiable image prior.

\section{Score-Based Priors}
\label{sec:score_based_priors}
We propose score-based priors as differentiable image priors that can be trained on any dataset and employed for principled inverse imaging. 
For example, one can learn a prior on face images by training a score model $\mathbf{s}_\theta(\mathbf{x},t)$ on CelebA~\cite{liu2015faceattributes}. Then one can model any posterior with a face-image prior by appealing to the function $\log p_\theta(\mathbf{x})$ (Fig.~\ref{fig:teaser}).

\subsection{Log-Probability Computation}
Our method leverages previous work that shows how to compute image probabilities under the SDE framework~\cite{song2021scorebased}. This feature has been mostly discussed in theory but is yet to be demonstrated in practice. In this section, we discuss log-probability computation under a score-based diffusion model and empirically validate it. 

\textbf{Probability flow ODE.}
Computing probabilities requires inverting the sampling process: the probability of an image $\mathbf{x}_0$ depends on the probability of the $\mathbf{x}_T$ that would have resulted in that image through reverse diffusion. There is an ordinary differential equation (\textit{probability flow ODE})~\cite{song2021scorebased} that makes both the forward and reverse SDEs (Eqs.~\ref{eq:forward}, \ref{eq:reverse}) invertible, inducing the same time-dependent probability distribution $p_t$ but without Brownian motion. The ODE is the same forward and backward in time, defining a bijective mapping between $p_t$ and $p_{t'}$ for any two times $t,t'\in[0,T]$.\footnote{The implicit assumption is that the score model is well-trained such that $p_0\approx p_\text{data}$, where $p_\text{data}$ is the true distribution of the training data.} For a score model $\mathbf{s}_\theta(\mathbf{x},t)\approx\nabla_{\mathbf{x}}\log p_t(\mathbf{x})$, the learned probability flow ODE is given by
\begin{align}
\label{eq:ode}
    \frac{\mathrm{d}\mathbf{x}_t}{\mathrm{d}t}&=\mathbf{f}(\mathbf{x}_t,t)-\frac{1}{2}g(t)^2\mathbf{s}_\theta(\mathbf{x}_t,t)=:\tilde{\mathbf{f}}_\theta(\mathbf{x}_t,t).
\end{align}

\textbf{Log-probability formula.} By the continuous-time change-of-variables formula~\cite{chen2018neural}, the log-probability of an image $\mathbf{x}=\mathbf{x}_0$ under the $p_0$ distribution is given by the log-probability of $\mathbf{x}_T$ under the $p_T$ Gaussian, plus a normalization factor accounting for the change in probability density from $\mathbf{x}_0$ to $\mathbf{x}_T$. We compute the log-probability under the learned ODE (Eq.~\ref{eq:ode}) by solving an initial-value problem:
\begin{align}
\label{eq:log_prob_formula}
    \log p_0(\mathbf{x}_0)&=\log p_T(\mathbf{x}_T)+\int_{0}^T \nabla\cdot\tilde{\mathbf{f}}_\theta(\mathbf{x}_t,t)\mathrm{d}t,
\end{align}
where $\mathbf{x}_0=\mathbf{x}$. The divergence $\nabla\cdot\tilde{\mathbf{f}}_\theta(\mathbf{x}_t,t)$ quantifies the instantaneous change in log-probability of $\mathbf{x}_t$ caused by applying $\tilde{\mathbf{f}}_\theta(\mathbf{x}_t,t)$ in either time direction. It can be estimated with Hutchinson-Skilling estimation of the trace of $\frac{\partial}{\partial \mathbf{x}_t}\tilde{\mathbf{f}}_\theta(\mathbf{x}_t,t)$~\cite{song2021scorebased}. 
We denote the log-probability function of a score-based prior as $\log p_\theta:=\log p_0$.

\subsection{Log-Probability Validation}
\begin{figure}[ht]
    \centering
    \includegraphics[width=0.4\textwidth]{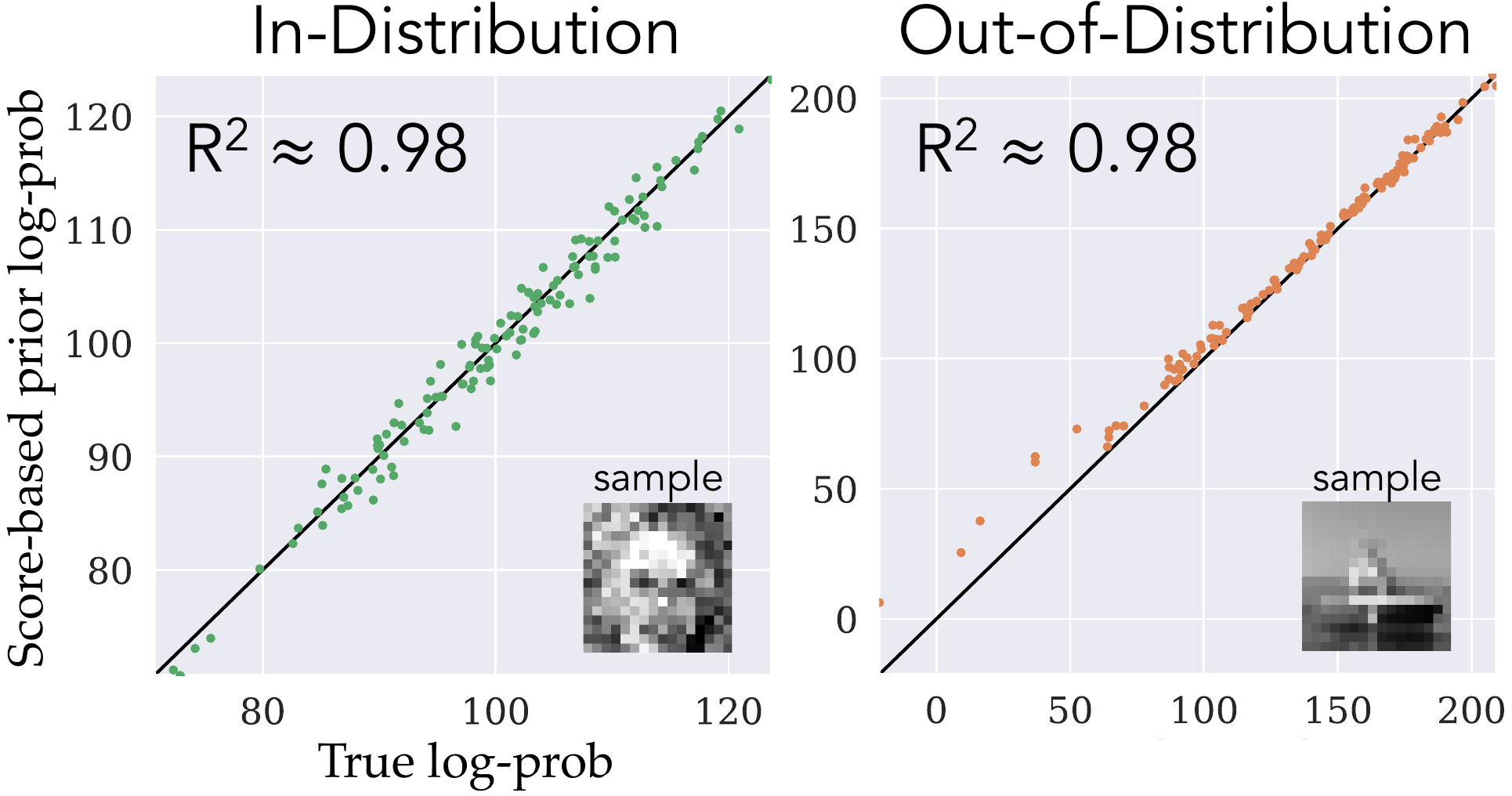}
    \caption{Log-probabilities of the score-based prior vs. ground-truth. Black line indicates perfect agreement. \textbf{In-Distribution.} The log-probabilities of 128 samples from the Gaussian ground-truth distribution were evaluated (shown as scatter points). Score-based log-probabilities are strongly correlated with ground-truth log-probabilities ($R^2\approx0.98$). \textbf{Out-of-Distribution.} The log-probabilities of test images from CIFAR-10 (scaled to $16\times 16$) are shown. The score-based prior generalizes well out of distribution.}
    \label{fig:gt_logprobs}
\end{figure}

\begin{figure}[t]
    \centering
    \includegraphics[width=0.3\textwidth]{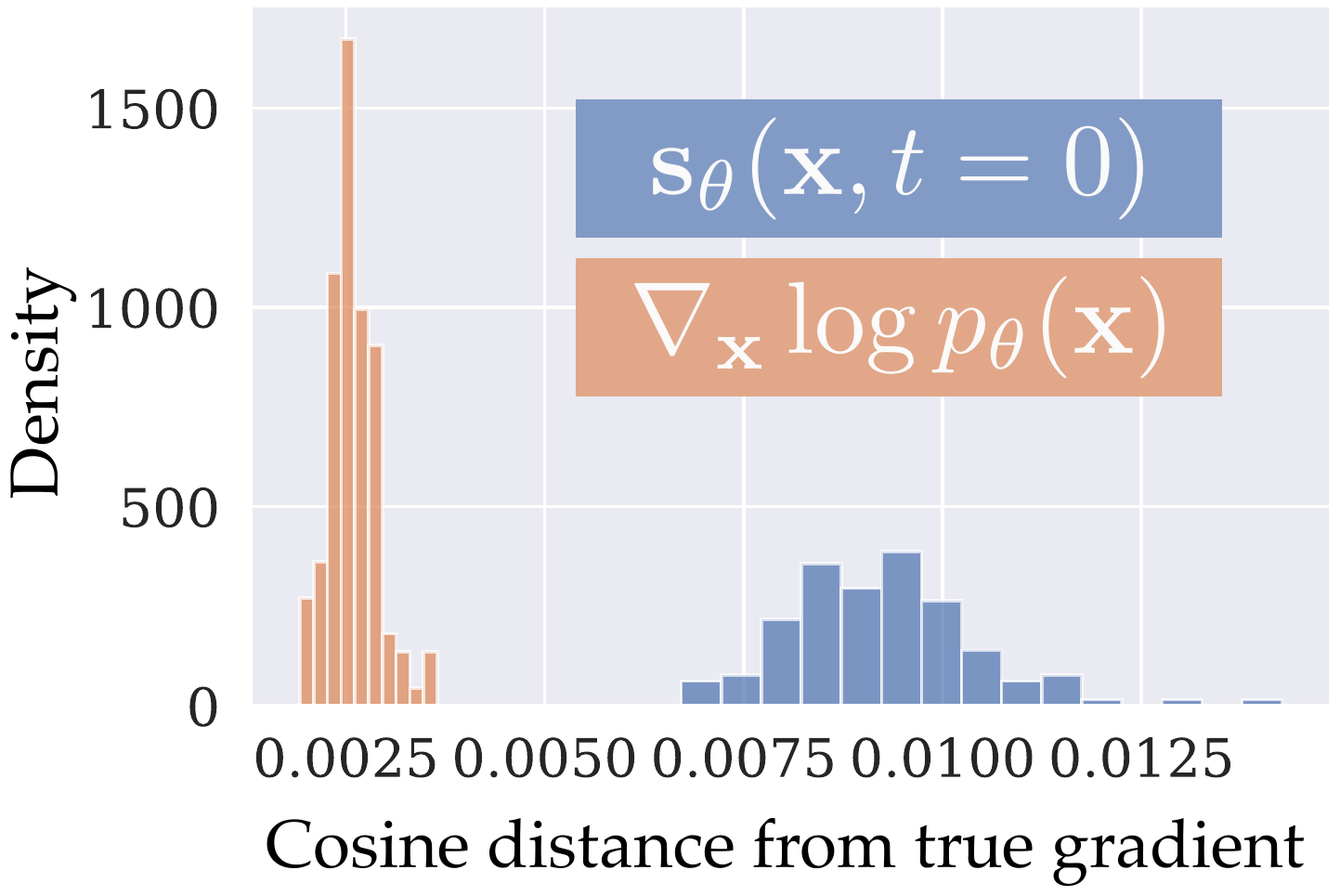}
    \caption{ODE gradients vs.~score-model outputs. Histogram shows density of cosine distance between the estimated gradient and true gradient for 128 samples from the ground-truth Gaussian. $\mathbf{s}_\theta(\mathbf{x},t=0)$ is the score-model-approximated gradient ($t=0.001$ is actually used for numerical stability). $\nabla_{\mathbf{x}}\log p_\theta(\mathbf{x})$ is computed numerically according to the probability flow ODE.
    }
    \label{fig:grad_cosine}
\end{figure}

In several experiments (Figs.~\ref{fig:gt_logprobs}, \ref{fig:grad_cosine}, \ref{fig:dpi_gt_deblurring}), we compare a learned score-based prior to a known ground-truth distribution. The ground-truth is a Gaussian distribution of $16\times 16$ grayscale images, whose mean and preconditioned covariance were fit to CelebA face images. The score model was trained on samples from this Gaussian. Fig.~\ref{fig:gt_logprobs} shows our empirical analysis of log-probabilities under the score-based prior versus the true prior.  We verify accuracy on both in-distribution and out-of-distribution images.

\textbf{Gradients.} As many Bayesian-inference approaches require a \textit{differentiable} prior, we validate that the gradient function $\nabla_{\mathbf{x}}\log p_\theta(\mathbf{x})$ provides more accurate gradients than the score model $\mathbf{s}_\theta(\mathbf{x},t=0)\approx\nabla_{\mathbf{x}}\log p_0(\mathbf{x})$. It is tempting to apply the score model as a cheap gradient approximator since it is designed in theory as such, but in practice, it does not generalize to out-of-distribution data.
It is more reliable to compute the gradient of the entire ODE solve of $\log p_\theta(\mathbf{x})$ with respect to $\mathbf{x}$. Fig.~\ref{fig:grad_cosine} shows that given samples from the ground-truth Gaussian, gradients computed according to the ODE are closer to the true gradients (in terms of cosine similarity) than score-model outputs. Fig.~\ref{fig:dpi_gt_deblurring} (appearing in Sec.~\ref{sec:posterior_sampling}) shows that using score-model outputs as gradients leads to an incorrect posterior.

\textbf{Implementation details.} Our code\footnote{Website: \scriptsize{\url{http://imaging.cms.caltech.edu/score_prior}}} is written in JAX and  Diffrax~\cite{kidger2022neural} to make it easy to just-in-time (JIT) compile log-probability computations, select ODE solvers, and compute autograds. For experiments in this paper, we used 5th-order Runge-Kutta solvers~\cite{dormand1980family,tsitouras2011runge} and Hutchinson-Skilling trace estimation~\cite{hutchinson1989stochastic,skilling1989eigenvalues}. We approximated gradients with the continuous adjoint method~\cite{kidger2021on}. The supplementary text includes further practical details about solvers, trace estimation, and our experimental setups.

\section{Posterior Sampling with Score-Based Priors}
\label{sec:posterior_sampling}
With score-based priors, we can solve inverse problems by sampling from rich posteriors given any image prior (assuming sufficient training data) and any measurements (assuming a known forward model). We accomplish posterior sampling by plugging the score-based prior into a variational-inference method. This both highlights the applicability of score-based priors to established optimization methods and provides a solution to the open problem of posterior sampling with unconditional diffusion models. In this section, we describe our chosen approach, \textit{Deep Probabilistic Imaging} (DPI), which converges faster than MCMC and provides efficient sampling~\cite{sun2021deep}.

\subsection{Main Idea for Posterior Sampling}
Our main idea for unlocking the rich prior of a diffusion model is to remove the concept of diffusion and consider the prior as a fixed distribution just like the likelihood.
This lets us directly model the posterior $\log p(\mathbf{x}|\mathbf{y})$.

\begin{figure}[htb]
    \centering
    \includegraphics[width=0.48\textwidth]{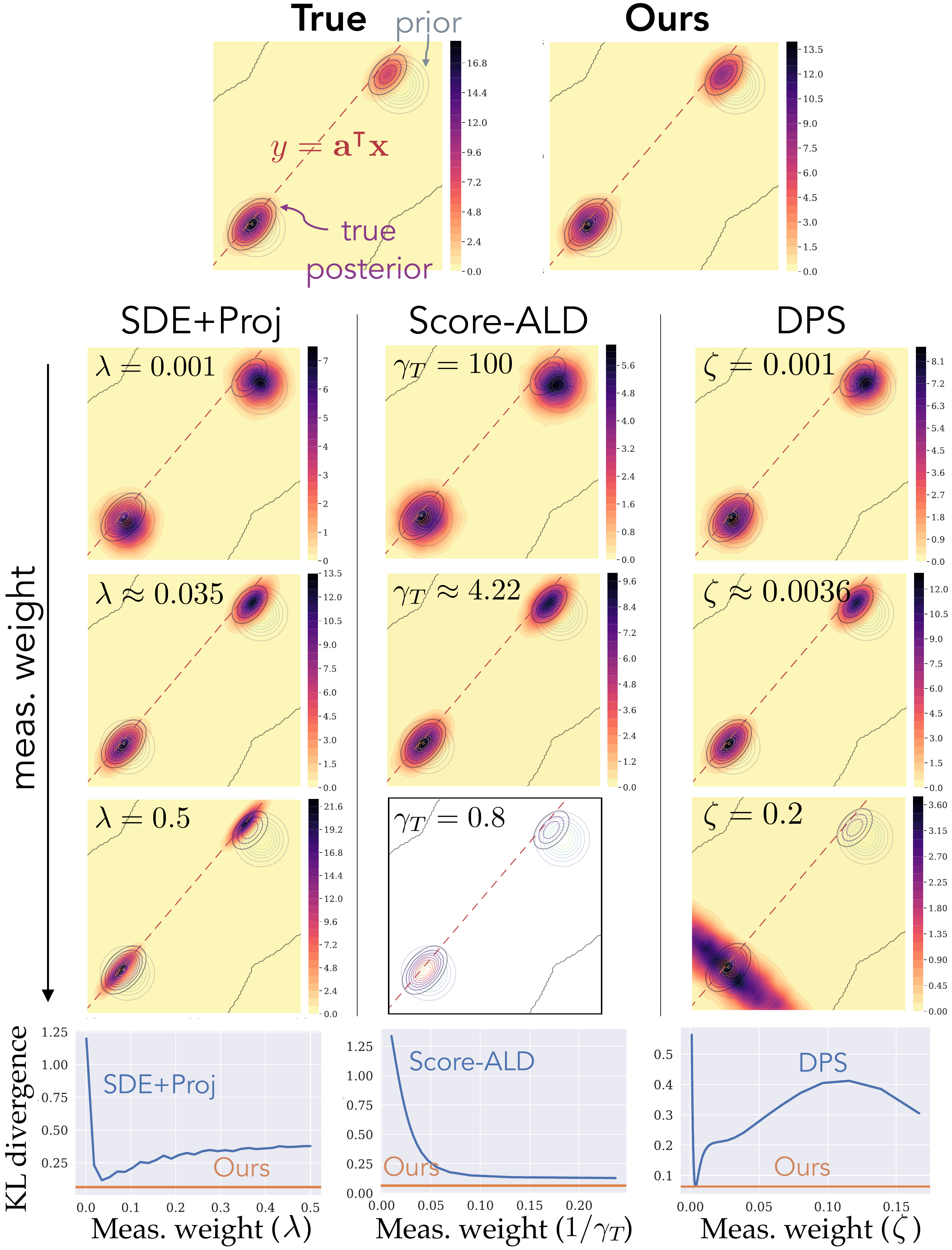}
    \caption{Baseline methods~\cite{song2022solving,jalal2021robust,chung2023diffusion} do not sample true posterior. Heatmap depicts $p(\mathbf{x}| y)$ approximated from samples; contour lines depict true posterior. All methods use the same (true) score function. For baselines, we did a grid search to find the optimal hyperparameter weight (ALD and DPS hyperparameters were distilled into a global hyperparameter).
    The \textbf{KL divergence} from the estimated posterior to the true posterior was approximated for each hyperparameter value. No matter the value, baselines do not get more accurate than our hyperparameter-free method.
    For instance, with ``Best meas.~weight'', all baselines sample from both modes equally, even though the lower-left mode should have more density.
    A poor hyperparameter setting can even lead to unstable sampling (see ``High meas.~weight'' of Score-ALD, where samples became NaNs due to slightly over-weighted measurements).
    }
    \label{fig:2d_example_bimodal}
\end{figure}

Previous methods treat the diffusion model as an implicit prior and entangle measurements with diffusion, which inhibits true posterior sampling. Posterior reverse diffusion requires the \textit{posterior} score function
$$
\nabla_{\mathbf{x}_t}\log p_t(\mathbf{x}_t\mid\mathbf{y})=\nabla_{\mathbf{x}_t}\log p_t(\mathbf{y}\mid\mathbf{x}_t)+\mathbf{s}_\theta(\mathbf{x},t)
$$
for all $t\in[0, T]$. The score model gives the prior score, but the likelihood score is only defined for $t=0$. For every diffusion time $t$, the true likelihood is defined by an intractable integral over all $\mathbf{x}_0$:
$$
p_t(\mathbf{y}\mid\mathbf{x}_t)=\int_{\mathbf{x}_0}p(\mathbf{y}\mid\mathbf{x}_0)p(\mathbf{x}_0\mid\mathbf{x}_t) \mathrm{d}\mathbf{x}_0.
$$
Previous methods either abandon the true measurement uncertainty~\cite{choi2021ilvr,chung2022improving,chung2022come,chung2022score,graikos2022diffusion,song2022solving,adam2022posterior} or strongly approximate $p_t(\mathbf{y}\mid\mathbf{x}_t)$~\cite{chung2023diffusion,jalal2021robust,kawar2022denoising,song2023pseudoinverseguided}.
This necessitates hyperparameter(s) for determining the importance of measurements versus the prior, making the estimated posterior more of a \textit{conditional} distribution rather than a principled posterior.
Different hyperparameter settings have drastic effects on the estimated posterior: if measurements are under-weighted, then the posterior is overly-biased toward the prior and may contain misleading data; if measurements are over-weighted, then the samples may collapse onto a subspace that does not make sense under the prior. Fig.~\ref{fig:2d_example_bimodal} shows these outcomes on a 2D example. It also shows that, even with ideally-tuned hyperparameters (which require knowledge of ground-truth), previous methods cannot capture the true posterior. We describe the baseline methods in Sec.~\ref{sec:posterior_sampling_results}.

\subsection{Selected Approach: Variational Inference with a Normalizing Flow}
Our approach for directly modeling the posterior is rooted in variational inference. Following the method proposed in DPI~\cite{sun2021deep,sun2022alpha}, we define a family of distributions $q_\phi$ via a RealNVP~\cite{dinh2016density} normalizing flow with parameters $\phi$, which we optimize to approximate the desired posterior. 
For a score-based prior with parameters $\theta$, the objective is
\begin{align}
\label{eq:dpi_objective}
    \phi^*&=\arg\min_\phi D_{\text{KL}}\left(q_\phi\Vert p_\theta(\cdot\mid\mathbf{y})\right) \\
    &=\arg\min_\phi \mathbb{E}_{\mathbf{x}\sim q_\phi}\left[-\log p(\mathbf{y}\mid\mathbf{x})-\log p_\theta(\mathbf{x})+\log q_\phi(\mathbf{x})\right]. \notag
\end{align}
This variational objective includes the log-posterior and an entropy term, $\mathbb{E}_{\mathbf{x}\sim q_\phi}[\log q_\phi(\mathbf{x})]$, which can be tractably computed under the RealNVP.\footnote{As discussed in Sec.~\ref{sec:related_work}, discrete normalizing flows allow for exact probabilities, but this feature is not reliable on out-of-distribution data. We therefore only use $\log q_\phi(\mathbf{x})$ during fitting and then once $\phi$ is fixed, only use the RealNVP for sampling.} During fitting, the expectation is Monte-Carlo approximated with a batch of samples from the RealNVP. Note that the \textbf{absence of hyperparameters} in this objective is a feature of the score-based prior (not DPI).
Because our prior is truly probabilistic, there is no need for a hand-tuned weight.

\begin{figure}[t]
    \centering
    \includegraphics[width=0.42\textwidth]{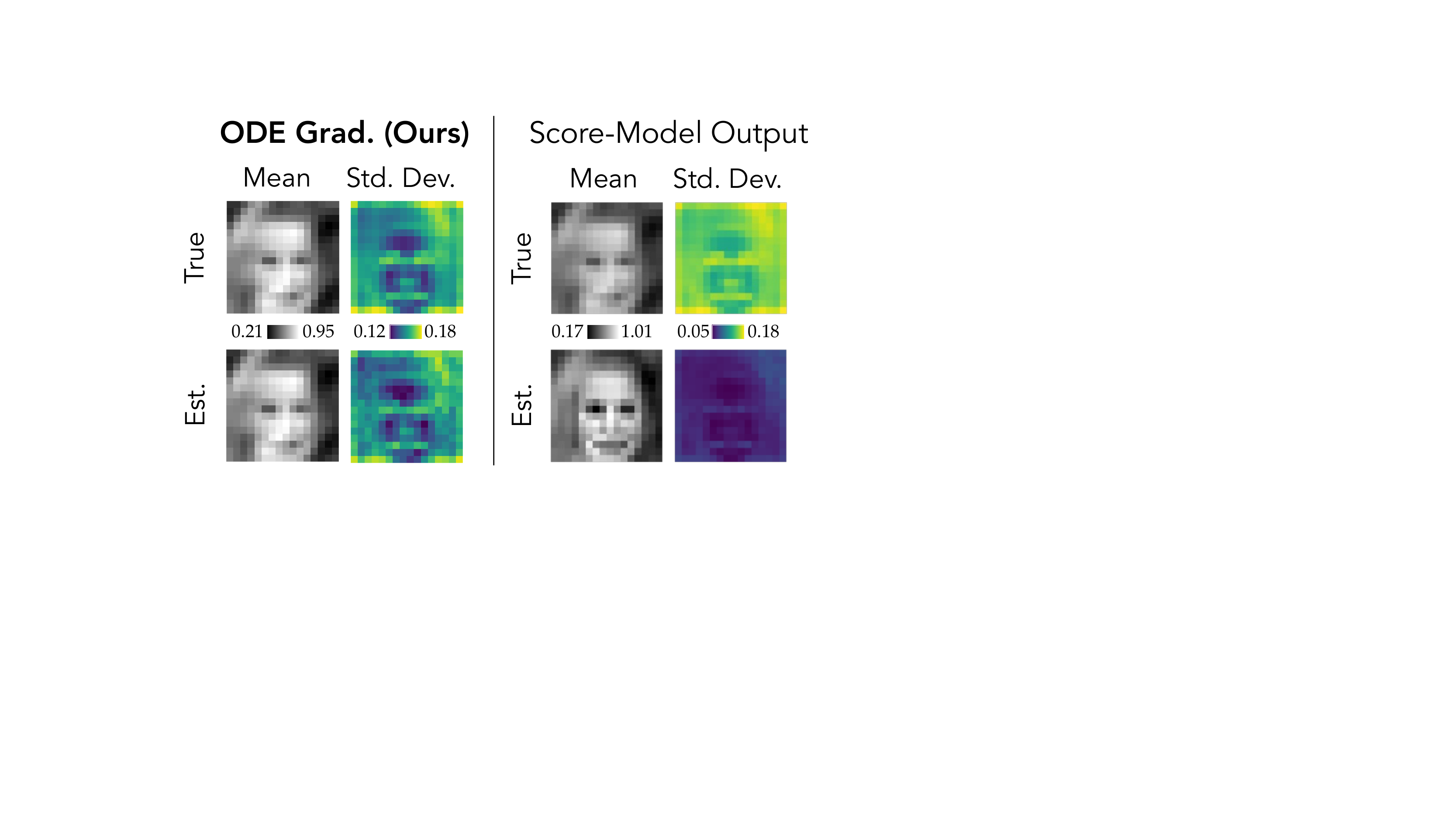}
    \caption{True vs.~estimated posterior. Measurements are the 6.25\% lowest DFT spatial-frequencies of an image from the prior and have i.i.d. noise with $\sigma=\lvert 1 \rvert$. The true mean and variance were derived analytically since the prior is the Gaussian ground-truth distribution, and the likelihood distribution is also Gaussian. The mean and variance were estimated from 10240 samples.}
    \label{fig:dpi_gt_deblurring}
\end{figure}

As evidence that our method correctly samples the posterior, Fig.~\ref{fig:dpi_gt_deblurring} compares an estimated posterior to ground-truth. The score-based prior was trained on the ground-truth Gaussian used in Sec.~\ref{sec:score_based_priors}, and the task was to deblur an image from the true prior. The estimated mean and covariance closely agree with the analytical mean and covariance. Additionally, we find that directly using the score-model output leads to an incorrect posterior.

We note that accurate posterior sampling is dependent on accurate log-probabilities in the first place. We verify that log-probabilities under a score-based prior well-approximate those under a ground-truth Gaussian prior in Fig.~\ref{fig:gt_logprobs}, but log-probabilities are difficult to validate for complex priors (which motivates our work).

\section{Posterior Sampling Results}
\label{sec:posterior_sampling_results}

\subsection{Baseline Methods}
We include comparisons to three previous methods for posterior sampling with an unconditional diffusion model.
Fig.~\ref{fig:2d_example_bimodal} compares methods for a 2D example.

\textbf{SDE+Proj} (Song et al. 2022)~\cite{song2022solving}. The image $\mathbf{x}_t$ is projected onto a measurement subspace at each $t$ throughout reverse diffusion. A hyperparameter $\lambda$ determines measurement weight. This approach is purported to give plausible solutions rather than exact posterior samples. It is restricted to compressed-sensing linear inverse problems where the forward matrix has fewer rows than columns.

\textbf{Score-based annealed Langevin dynamics (Score-ALD)} (Jalal \& Arvinte et al. 2021)~\cite{jalal2021robust}. The authors propose ALD with approximate posterior scores:
{
\begin{align}
    \nabla_{\mathbf{x}_t}\log p_t(\mathbf{x}_t\mid\mathbf{y})\approx \mathbf{s}_\theta(\mathbf{x}_t,t) + \frac{\mathbf{A}^H(\mathbf{y}-\mathbf{Ax}_t)}{\sigma^2+\gamma_t^2},
\end{align}
}
where $\frac{\mathbf{A}^H(\mathbf{y}-\mathbf{Ax}_t)}{\sigma^2+\gamma_t^2}$ is the log-likelihood gradient assuming linear measurements ($\mathbf{y}=\mathbf{Ax}+\epsilon$) and Gaussian noise ($\epsilon\sim\mathcal{N}(\mathbf{0}_M,\sigma^2\mathbf{I}_M)$). $\gamma_t$ is a hyperparameter for the weight of the log-likelihood term at step $t$, thus calling for a hand-tuned approximation of the likelihood score at each $t$. One way to automatically adjust the magnitude of the likelihood score is to renormalize it to have the same magnitude as the prior score, which is what the authors do in their experiments. Without this trick, we find image posterior sampling highly sensitive to the $\gamma_t$ annealing schedule.

\textbf{Diffusion posterior sampling (DPS)} (Chung \& Kim et al. 2023)~\cite{chung2023diffusion}. The authors propose an approximation of $\log p_0(\mathbf{y}\mid\mathbf{x}_t)$ throughout reverse diffusion. A hyperparamter $\zeta_t$ determines the magnitude of the log-likelihood gradient at each $t$. Similar to ALD, the authors use a trick to determine the magnitude of the likelihood score, setting $\zeta_t:=\zeta/\lVert \mathbf{y}-\mathbf{f}(\hat{\mathbf{x}}_0(\mathbf{x}_t))\rVert$, where $\mathbf{f}$ is the measurement forward operator, and $\mathbf{x}_0(\mathbf{x}_t):=\mathbb{E}[\mathbf{x}_0\mid\mathbf{x}_t]$.

\subsection{Imaging Inverse Problems}
\begin{figure*}[t]
    \centering
    \includegraphics[width=0.85\textwidth]{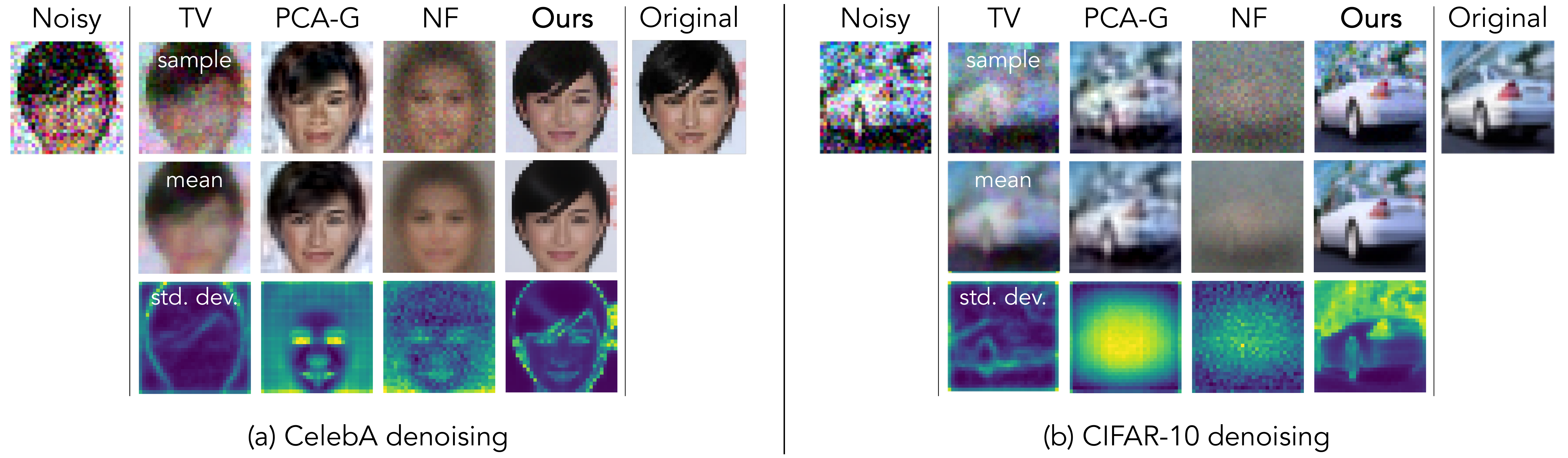}
    \caption{Denoising with different priors. The data-driven priors (PCA-G, NF, Ours) were trained on the CelebA training set in (a) and the CIFAR-10 training set in (b). \textbf{TV} regularization weight is 0.00025. \textbf{PCA-G} is a Gaussian based on the top 512 principal components of the training data. \textbf{NF} is a RealNVP with 64 affine-couple layers. TV, PCA-G, and NF are relatively simple priors that do not give rich posteriors. Score-based priors (\textbf{Ours}) capture complex spatial correlations (see sample quality and data-driven std. dev. maps).}
    \label{fig:denoising}
\end{figure*}
We demonstrate image posterior sampling on denoising, a version of deblurring, and interferometric imaging of a simulated black hole. There are two comparisons to perform: (1) score-based priors vs.~other \textit{priors} and (2) our posterior-sampling approach vs.~other \textit{posterior-sampling approaches} using an unconditional score model. (1) is done with a denoising experiment, and (2) is done with a deblurring experiment, although the findings are not task-dependent. We then focus on black-hole imaging as an endeavor that could benefit from score-based priors as tools in the scientific process.

\subsubsection{Validation of prior (denoising)}
We denoise images corrupted by i.i.d.~Gaussian noise with standard deviation $\sigma=0.2$ (20\% of the dynamic range). Fig.~\ref{fig:denoising} shows denoised posteriors given different priors: (1) \textbf{TV} regularization, (2) PCA-Gaussian (\textbf{PCA-G}), (3) RealNVP normalizing flow (\textbf{NF}), and (4) our score-based prior. DPI was used to approximate the posterior for each prior except PCA-G, which has an analytical Gaussian posterior.

Score-based priors provide more informative posteriors than traditional priors do. This is reflected in sample quality: as Tab.~\ref{tab:denoising} shows, the average SSIM and PSNR of posterior samples are highest when using a score-based prior. Score-based priors also provide richer posteriors: as shown by the empirical standard deviations in Fig.~\ref{fig:denoising}, our priors result in full posteriors with a data-driven uncertainty (e.g., uncertainty is higher on facial features like the eyes, ears, and hair of the CelebA image, whereas other priors are more unaware of facial structure). The RealNVP NF performed poorly as an image prior, perhaps because it struggles to generalize to non-training images and thus leads to unstable optimization of the variational posterior.
Please refer to the supplementary text for more details about this experiment.

\begin{table}[htb]
\resizebox{0.48\textwidth}{!}{
\begin{tabular}{|r||c|c|c|c|c|}
\hline
\diagbox{Image}{Prior}
       & \begin{tabular}[c]{@{}c@{}}CelebA\\ (Ours)\end{tabular}         & \begin{tabular}[c]{@{}c@{}}CIFAR\\ (Ours)\end{tabular} & \begin{tabular}[c]{@{}c@{}}Same dataset\\ (PCA-G)\end{tabular} & \begin{tabular}[c]{@{}c@{}}Same dataset\\ (NF)\end{tabular} & TV                                                  \\ \hline\hline
CelebA & \begin{tabular}[c]{@{}c@{}}SSIM: \textbf{0.88}\\ PSNR: \textbf{25.1}\end{tabular} & \begin{tabular}[c]{@{}c@{}}0.71\\ 21.6\end{tabular}    & \begin{tabular}[c]{@{}c@{}}0.76\\ 20.5\end{tabular}      & \begin{tabular}[c]{@{}c@{}}0.31\\ 11.3\end{tabular}     & \begin{tabular}[c]{@{}c@{}}0.53\\ 17.4\end{tabular} \\ \hline
CIFAR  & \begin{tabular}[c]{@{}c@{}}SSIM: 0.79\\ PSNR: 19.0\end{tabular} & \begin{tabular}[c]{@{}c@{}}\textbf{0.85}\\ \textbf{21.0}\end{tabular}    & \begin{tabular}[c]{@{}c@{}}0.74\\ 19.4\end{tabular}      & \begin{tabular}[c]{@{}c@{}}0.22\\ 13.4\end{tabular}     & \begin{tabular}[c]{@{}c@{}}0.57\\ 17.6\end{tabular} \\ \hline
\end{tabular}
}
\caption{Denoising metrics. Row refers to test image in Fig.~\ref{fig:denoising}. Column refers to prior used for denoising.
Average SSIM and PSNR were computed across 128 posterior samples for each (test image, prior) pair.
The ``correct'' score-based prior (Ours) performs best for each test image, while even the ``incorrect'' score-based prior performs well compared to the other priors.
}
\label{tab:denoising}
\end{table}

\subsubsection{Validation of posterior sampling (deblurring)}
\begin{figure}[tb]
    \centering
    \includegraphics[width=0.5\textwidth]{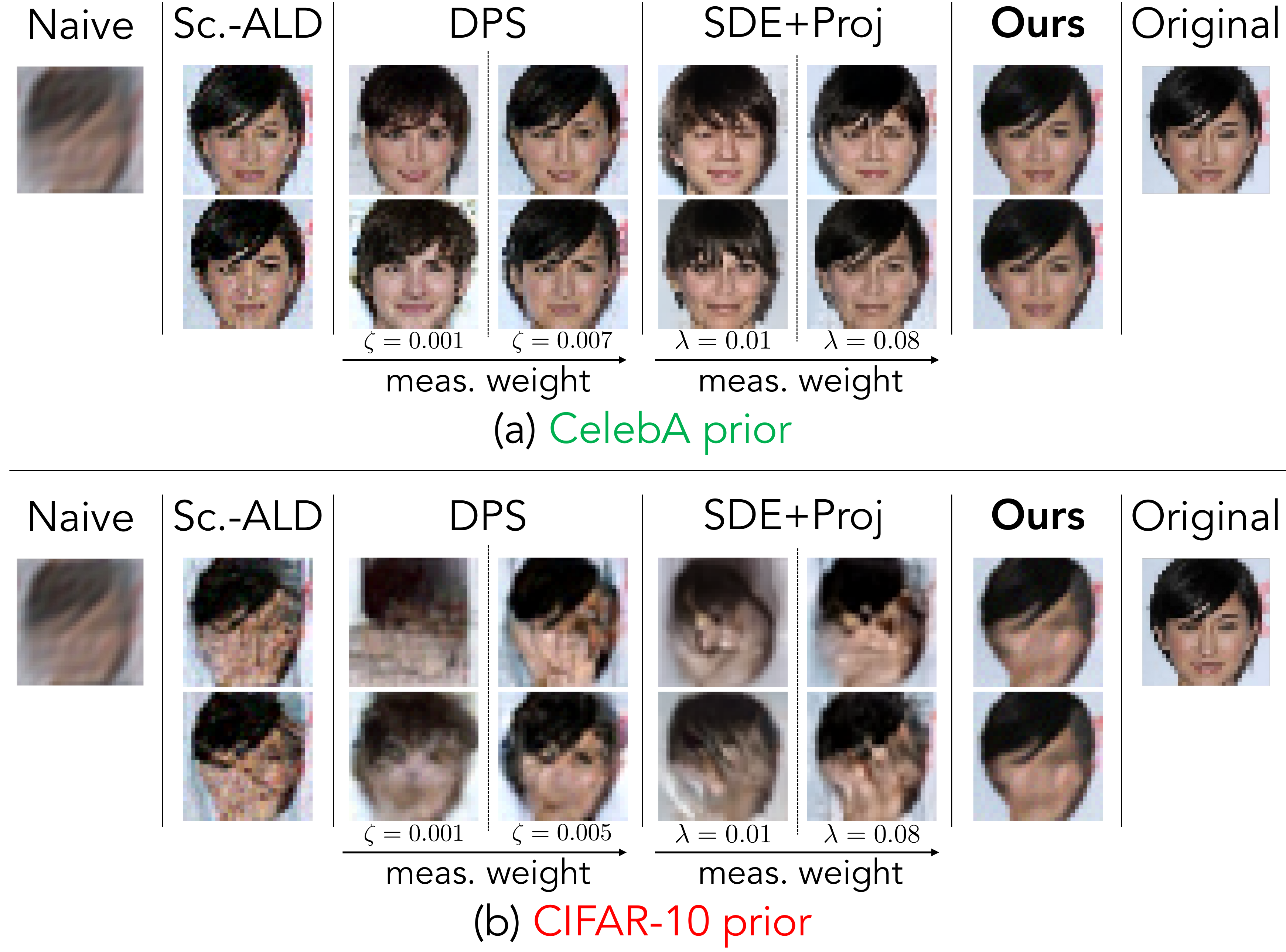}
    \caption{Deblurring (CelebA). Two samples from each method are shown. The source is a CelebA test image (Original). Score-ALD uses the likelihood-gradient renormalization trick. \textbf{(a) Using the ``correct'' prior.} All methods recover plausible posterior samples when the prior includes the true source image. With baselines, posterior variance depends on the meas.~weight, which is difficult to set without knowing the ground-truth posterior. \textbf{(b) Using an ``incorrect'' prior.} All methods (expectedly) are not able to recover a face. But baseline methods introduce heavy artifacts, while samples from Ours still look natural.}
    \label{fig:deblurring_celeba}
\end{figure}

\begin{figure}[htb]
    \centering
    \includegraphics[width=0.495\textwidth]{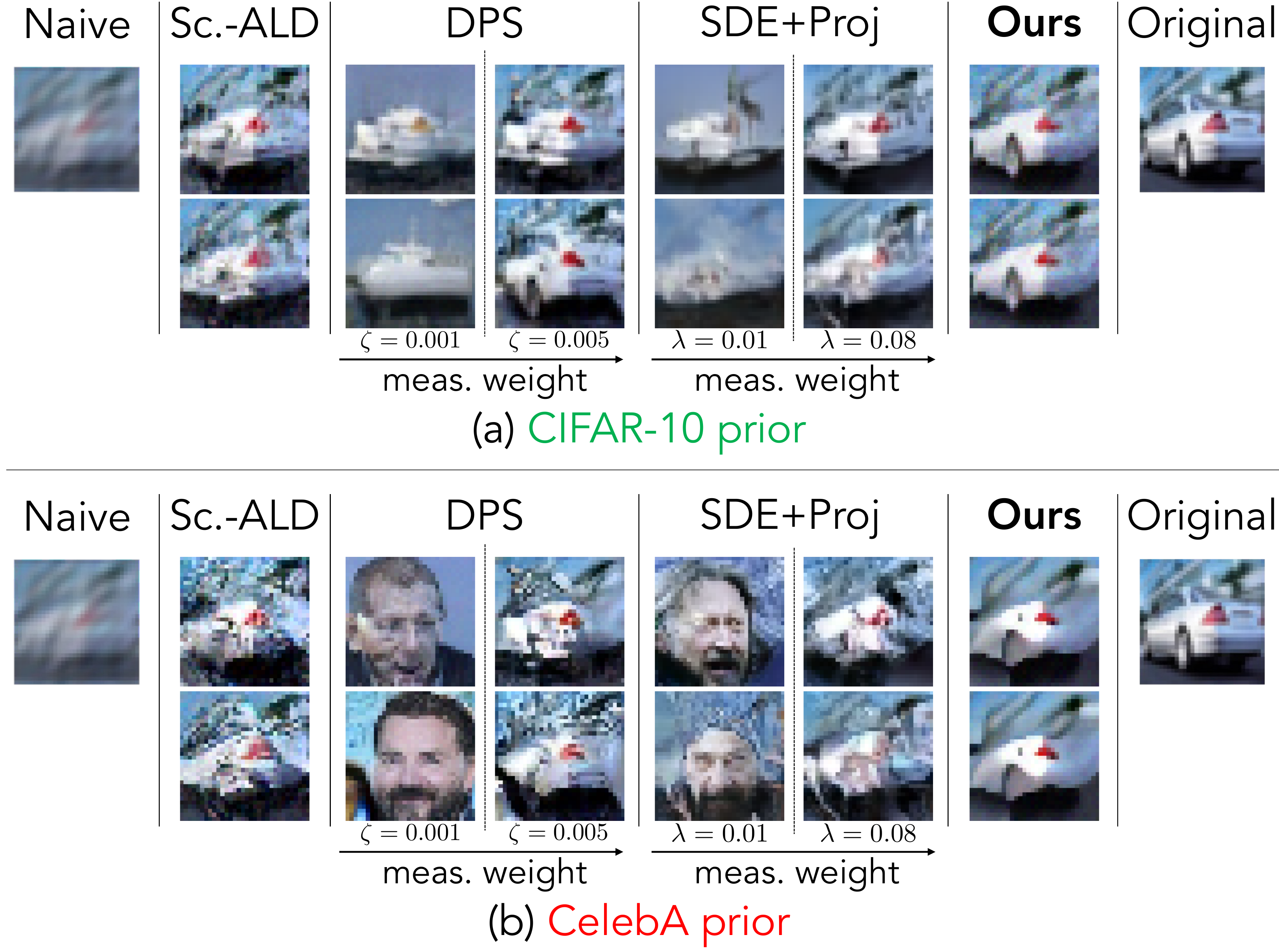}
    \caption{Deblurring (CIFAR-10). \textbf{(a) Using the ``correct'' prior.} For DPS and SDE+Proj, both weights achieve plausible samples, but variance differs drastically. \textbf{(b) Using an ``incorrect'' prior.} ALD, DPS, and SDE+Proj with high meas.~weight struggle to produce natural images when faced with out-of-distribution measurements. With lower meas.~weight, DPS and SDE+Proj hallucinate faces. This is also troubling since the posterior should not lie inside the prior in this case.}
    \label{fig:deblurring_cifar10}
\end{figure}

We consider the task of reconstructing an image from measurements of the lowest DFT spatial frequencies, which we call ``deblurring'' to simplify terminology. In our experiments, we observed the lowest 6.25\% DFT spatial frequencies with complex-valued measurement noise with standard deviation $\lvert\sigma\rvert=1.0$ ($\sim$0.2\% of the magnitude of the zero-frequency component).
Figs.~\ref{fig:deblurring_celeba}, ~\ref{fig:deblurring_cifar10} show results for a CelebA and CIFAR-10 source image, respectively, comparing our method, Score-ALD, DPS, and SDE+Proj. We used the same two score models (one trained on CelebA and one trained on CIFAR-10) for all methods. Our method outperforms others in terms of MSE, PSNR, and SSIM (e.g., in Fig.~\ref{fig:deblurring_celeba}(a), our posterior samples have an average PSNR of 24.75, while DPS, the best-performing method in terms of PSNR, achieved an average of 20.37). Full metrics for Figs.~\ref{fig:deblurring_celeba}, \ref{fig:deblurring_cifar10} are provided in the supplementary text.

Our posterior-sampling method is much more robust to mismatched priors than baselines are. Consider the CelebA prior applied to measurements of a CIFAR-10 image in Fig.~\ref{fig:deblurring_cifar10}(b). When the measurement weight in baselines is lower, they hallucinate faces, resulting in a posterior that lies within the prior. When it is higher, they introduce unnatural artifacts to fit the measurements.
Hyperparameter-dependent methods make it easy to mistakenly over-bias the prior or the measurements.
Without access to a ground-truth, one cannot reliably interpret a posterior if it is not the result of principled Bayesian inference.


\subsubsection{Interferometric imaging}
The scientific venture of black-hole imaging calls for principled inference with priors that are minimally hand-tuned.
Radio interferometry is a technique for imaging astronomical targets with high angular resolution by using a distributed array of radio telescopes. An interferometer collects sparse spatial-frequency measurements of the sky's image.
The Event Horizon Telescope (EHT) notably used this technique to take the first image of a black hole~\cite{event2019first}. In this work, we simulated interferometric measurements from the EHT telescope array using the \texttt{ehtim} package~\cite{chael2018interferometric}.\footnote{These simulated measurements contain thermal noise but exclude realistic atmospheric noise that results in additional phase corruption.}

\begin{figure}[htb]
    \centering
    \includegraphics[width=0.45\textwidth]{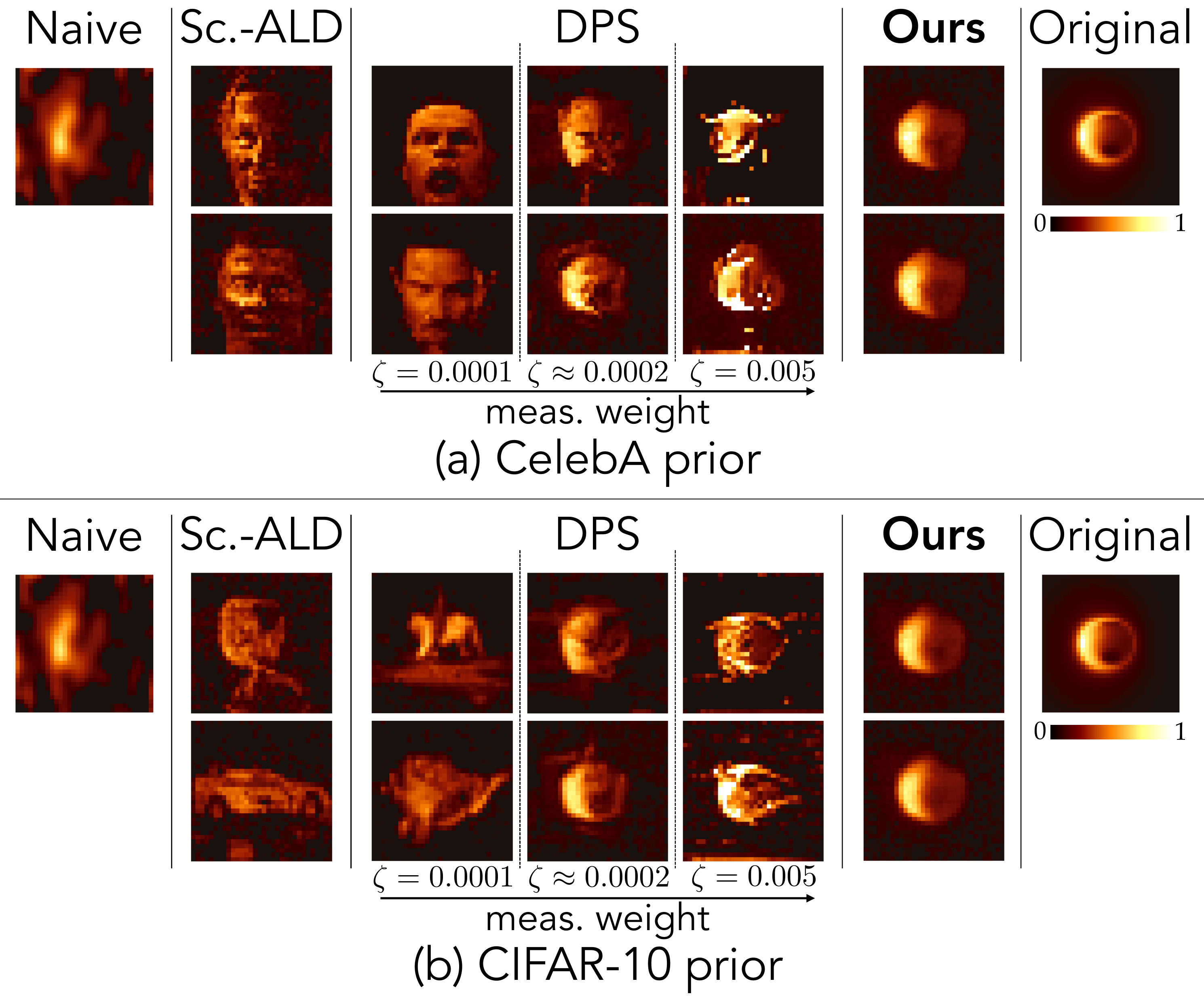}
    \caption{Interferometric imaging of a synthetic black-hole image. Two random samples from each method are shown (images are all shown with the same color scale).
    Both baselines struggle to balance measurements and prior, hallucinating faces in (a) and a car and horse in (b). Posterior samples become unstable as meas.~weight increases, as evidenced by the unnatural structures resulting from the highest meas.~weight in DPS. Regardless of meas.~weight, variance of baseline samples far exceeds ours (ALD  max. std. dev. is 0.202; DPS max. std. devs. are 0.237, 0.242, 0.444, resp., as meas.~weight increases; our max. std. dev. is 0.036). \textbf{Ours} produces samples that agree with the true structure, automatically balancing likelihood and prior.}
    \label{fig:eht}
\end{figure}

The first image of a black hole, while the result of carefully-obtained measurements from the EHT array, was only possible with image priors (technically formulated as regularizers). EHT scientists handcrafted many priors, each bringing different biases to the image reconstruction. Only the common structure found between these priors could be reliably interpreted, such as the diameter of the photon ring surrounding the black hole~\cite{event2019first}. Score-based priors could streamline this process as principled, hyperparameter-free priors that can be easily trained on different image distributions. Plugged into the imaging algorithm, they provide a collection of posteriors that incorporate different image statistics while maintaining measurement consistency.

It is important to remember that a posterior exists for \textit{any} combination of a prior and measurements, no matter how far the prior is from the source image. Faithfully modeling the posterior that arises from a given prior is especially crucial in a task like black-hole imaging. Since it is impossible to train a prior on real black holes, any data-driven prior likely does not perfectly agree with the source image.
Similar to our deblurring experiments (Figs.~\ref{fig:deblurring_celeba}, \ref{fig:deblurring_cifar10}), Fig.~\ref{fig:eht} highlights that score-based priors are robust to mismatches between the prior and true underlying distribution. Fig.~\ref{fig:eht} shows results using score-based priors trained on CelebA and CIFAR-10. The simulated measurements, from all five telescopes in the array, contain enough information for both priors to recover the underlying image structure. We compare our results to those of Score-ALD and DPS (SDE+Proj does not support this type of forward model).~\footnote{The EHT forward matrix has more rows than columns, but it is ill-conditioned because measurements are highly correlated.} Unlike these methods, which introduce either prior-related features (e.g., face) or unstructured artifacts (e.g., random pixels to agree with measurements), our hyperparameter-free method automatically balances the measurements and prior.

As shown in Fig.~\ref{fig:teaser}, a score-based prior visibly affects the posterior wherever measurements are not sufficient to constrain image structure. For instance, when applying a score-based prior trained on CelebA, as measurements are removed, more facial features appear in the posterior images. Given enough measurements, the recovered structure in the posterior samples can be reliably analyzed for scientific interpretation. With our framework, it is also possible to train a collection of score-based priors and look for common features between the posteriors that arise from the different priors. These common features are more likely to reflect the true underlying image.

\section{Limitations}
A score-based prior is a learned approximation of a desired image distribution, with its correctness tied to the correctness of the score model.
There is also a tradeoff between the complexity of the prior and compute time/memory. To obtain exact log-probabilities with a score-based prior, it is necessary to solve an ODE (often repeatedly throughout an optimization algorithm). Theoretical and practical improvements can be made to reduce this cost, such as making the score model more generalizable to quickly estimate $\nabla_\mathbf{x}\log p_\theta(\mathbf{x})$.
We also note that our approach for posterior sampling approximates the true posterior with a parameterized variational distribution, whose expressiveness determines the quality of the approximation.

\section{Conclusion}
We have shown how to turn a score-based diffusion model into a principled prior, specifically demonstrating a variational approach for posterior sampling. Results on 2D data establish the soundness of the approach, while our image denoising and deblurring experiments show the modularity of the score-based prior (applying the same prior across multiple tasks without any hand-tuning), its expressiveness (comparing to traditional priors), and its robustness (comparing to diffusion-based approaches to inverse problems). Score-based priors are especially useful for scientific imaging, such as interferometric imaging of black holes.
Our work opens a new direction in computational imaging that merges data-driven priors and principled inference.

\section*{Acknowledgments}
The authors would like to thank Michael Brenner for his helpful discussions throughout the project. They thank Yang Song, Zelda Mariet, Tianwei Yin, Patrick Kidger, and Mauricio Delbracio for their insightful feedback. Thanks also to Aviad Levis for his help with EHT software and Patrick Kidger for his help with Diffrax. BTF and KLB acknowledge funding from NSF Awards 2048237 and 1935980 and the Amazon AI4Science Partnership Discovery Grant. BTF is supported by the NSF GRFP.

{\small
\bibliographystyle{ieee_fullname}
\bibliography{main}
}
\end{document}


\title{Score-Based Diffusion Models as Principled Priors for Inverse Imaging:\\Supplemental}

\maketitle
\hypersetup{linkcolor=blue}
\tableofcontents
\hypersetup{linkcolor=red}
\appendix

\section{Implementation Details}
In this section, we discuss practical considerations for numerically computing log-probabilities and gradients under a score-based prior. We also discuss the experimental setups of our presented results.
\ificcvfinal\thispagestyle{empty}\fi

\subsection{Score-Based Priors}
\subsubsection{Log-probability computation}
Recall that for a pretrained score model $\mathbf{s}_\theta(\mathbf{x},t)$, the log-probability formula is given by
\begin{align}
\label{eq:log_prob_formula}
    \log p_0(\mathbf{x}_0)&=\log p_T(\mathbf{x}_T)+\int_{0}^T \nabla\cdot\tilde{\mathbf{f}}(\mathbf{x}_t,t;\theta)dt,
\end{align}
where $\tilde{\mathbf{f}}(\mathbf{x}_t,t;\theta)$ comes from the probability flow ODE:
\begin{align}
\label{eq:ode}
    d\mathbf{x}_t=\left[\mathbf{f}(\mathbf{x}_t,t)-\frac{1}{2}g(t)^2\mathbf{s}_\theta(\mathbf{x}_t,t)\right]dt=:\tilde{\mathbf{f}}(\mathbf{x}_t,t)dt.
\end{align}
Given an image $\mathbf{x}$, to compute $\log p_\theta(\mathbf{x})$ under the score-based prior parameterized by $\theta$, we have to solve an initial-value problem, where $\mathbf{x}_0=\mathbf{x}$ and $\frac{d\mathbf{x}_t}{dt}=\tilde{\mathbf{f}}(\mathbf{x}_t,t;\theta)$.

\textbf{Log-probability estimation.} The two implementation decisions that most affect log-probability accuracy are: (1) which ODE solver to use and (2) how to estimate the divergence in Eq.~\ref{eq:log_prob_formula}. To deal with (1), our code uses Diffrax~\cite{kidger2022neural}, a JAX library for differential equations, to easily swap out solvers and adaptively select time steps. As for (2), we use Hutchinson-Skilling estimation with multiple trace estimators to reduce the variance of log-probability and gradient calculations.

\textbf{ODE solver.} Tab.~\ref{tab:solvers} shows how different solvers affect time-efficiency and KL divergence to a ground-truth distribution. The ground-truth is the Gaussian distribution used in Main Sec.~3.2. This suggests that Bogacki--Shampine's 3/2 method and Dormand-Prince's 5/4 method offer a good balance between efficiency and accuracy. Note, however, that score-based priors trained on different datasets may show different trends. It is always a good idea to evaluate the runtime of different solvers for a given score-based prior to find the most efficient solver.

\begin{table}[!h]
\label{tab:solvers}
\begin{center}
\begin{tabular}{||c|c|c||} 
 \hline
 Solver & $D_{\text{KL}}(q\lVert p)$ ($\downarrow$) & NFE low. bound ($\downarrow$) \\
 \hline\hline
 Euler* (1st order) & 0.848 & 4092 \\ 
 \hline
 Heun (2nd order) & 0.478 & 312 \\
 \hline
 Bosh3 (3rd order) & 0.453 & 81 \\
 \hline
 Tsit5 (5th order) & 0.521 & 255 \\
 \hline
 Dopri5 (5th order) & 0.284 & 65 \\ 
 \hline
 Dopri8 (8th order) & 0.422 & 1440 \\
 \hline
\end{tabular}
\caption{KL divergence depending on the solver used for log-probability computation. ``Euler'' used a fixed step-size of $1/4092$. All other solvers used adaptive step-sizing, with the number of function evaluations (``NFE'') calculated as the number of solver steps times the order of the solver. The KL divergence was estimated from 512 samples from the ODE sampler.
}
\end{center}
\end{table}

\begin{figure}[htb]
    \centering
    \includegraphics[width=0.5\textwidth]{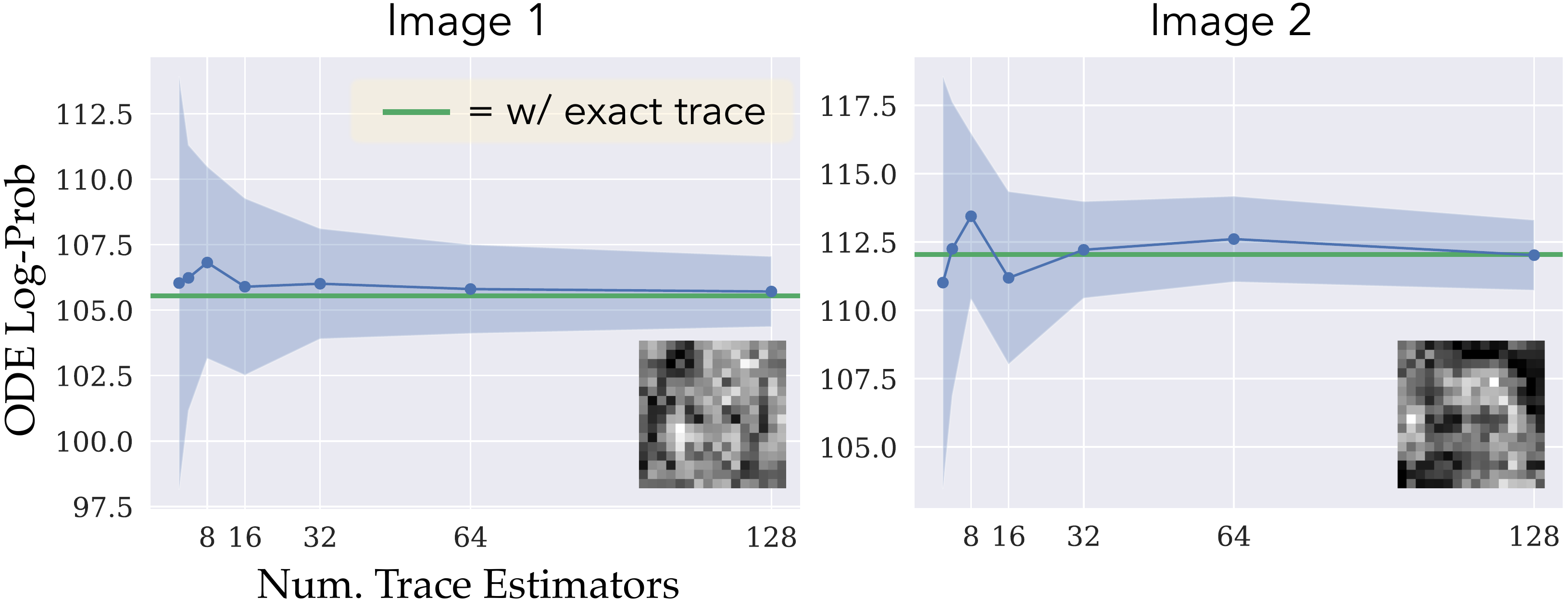}
    \caption{Mean and variance of log-probability values vs. number of trace estimators. The score-based prior was fit to the ground-truth Gaussian distribution used in Main Sec.~3.2. For each number of trace estimators (1, 2, 4, 8, 16, 32, 64, or 128), 50 trials of log-probability estimation were done with different random seeds (using the Dopri5 solver with adaptive step-sizing). The solid blue line indicates the mean of these trials, and the shaded region indicates one std. dev. above and below the mean. The solid green line shows the value resulting from exact trace calculation. The evaluated image is inset. As more trace estimators are used, the variance of the log-probability decreases.}
    \label{fig:trace_estimation_density}
\end{figure}

\begin{figure*}[t]
    \centering
    \includegraphics[width=0.6\textwidth]{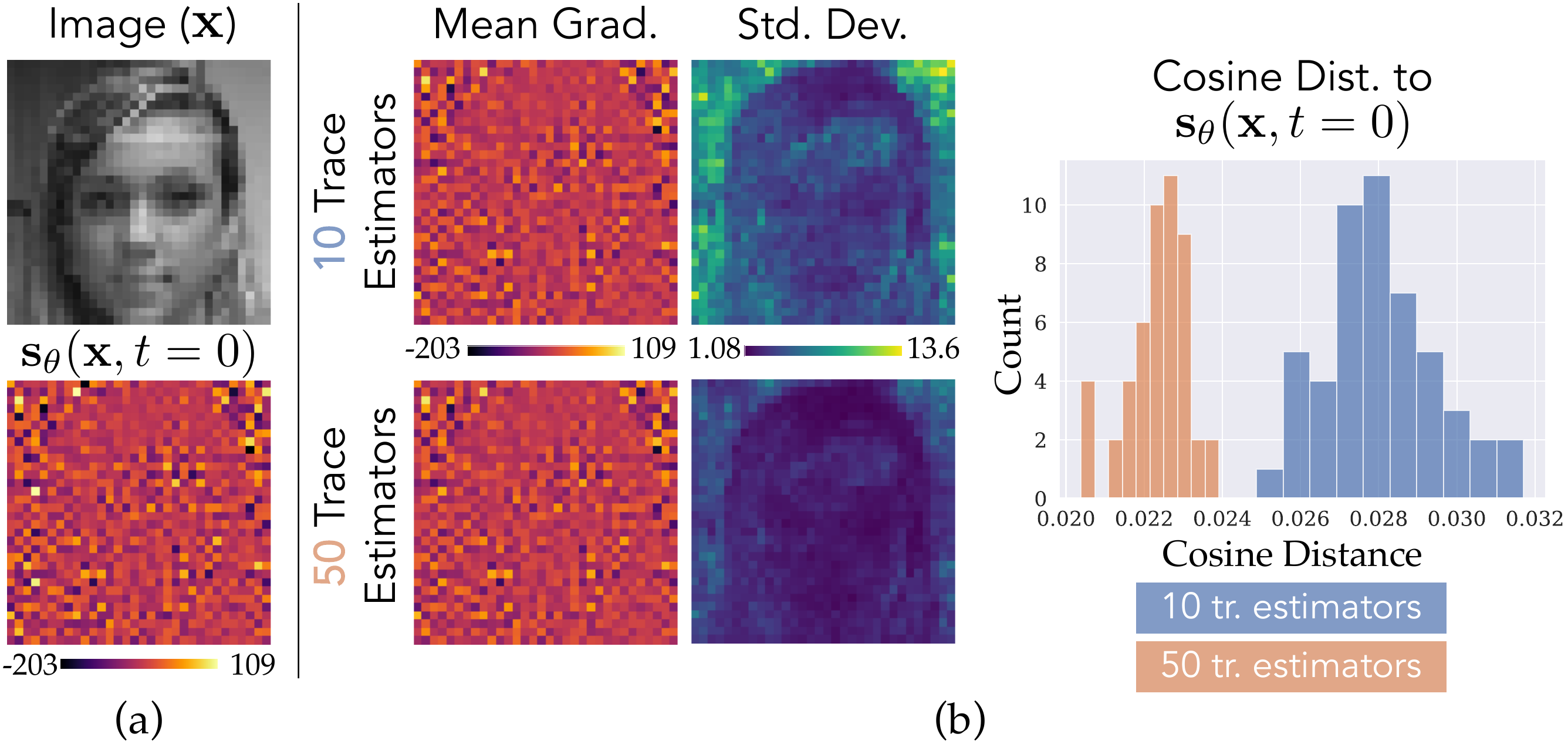}
    \caption{Mean and variance of $\nabla_{\mathbf{x}}\log p_\theta(\mathbf{x})$ with 10 vs.~50 trace estimators. The score-based prior was trained on $32\times 32$ grayscale CelebA images. \textbf{(a)} Test image $\mathbf{x}$ and gradient according to the learned score model, $\mathbf{s}_\theta(\mathbf{x},t)$, evaluated at $t= 0$. (In reality, we set $t=10^{-3}$ for numerical stability and perturbed $\mathbf{x}$ with noise accordingly.) Since the test image was drawn as $\mathbf{x}\sim p_0$, the score-model output should equal the true $\nabla_{\mathbf{x}}\log p_\theta(\mathbf{x})$. \textbf{(b)} Results of estimating  the gradient $\nabla_{\mathbf{x}}\log p(\mathbf{x})$ with the probability flow ODE including trace estimation. For both ``10 trace estimators'' and ``50 trace estimators'', 50 trials of gradient estimation were done with the continuous adjoint method. ``Mean Grad.'' and ``Std. Dev.'' are the average gradient and std. dev. of the gradient of all these runs. ``Cosine Dist. to $\mathbf{s}_\theta(\mathbf{x},t=0)$'' shows the histogram of the cosine distance between each gradient estimate and the score-model output, which we consider to be ground-truth. The results in (b) are evidence that trace estimation gives a good approximation of the gradient in expectation, but using fewer trace estimates causes higher variance. With 10 trace estimates, the median relative std. dev.~of the gradient is 16\%. With 50 trace estimates, it is 8.6\%. (Relative std. dev.~is computed as $\lvert \sigma \rvert / \lvert \mu \rvert$.) Also note that regions of highest variance are in the image background.}
    \label{fig:trace_estimation_grad}
\end{figure*}

\textbf{Trace estimation.} For high-dimensional data, trace estimation is necessary to estimate the divergence in Eq.~\ref{eq:log_prob_formula}. This causes variance in the estimated log-probabilities and gradients. Song et al.~\cite{song2021scorebased} use Hutchinson-Skilling with one trace estimator, but we use multiple trace estimators to reduce variance. In our implementation, the same trace estimators are applied to each image in a batch. Figs.~\ref{fig:trace_estimation_density} and \ref{fig:trace_estimation_grad} show the variance of densities and gradients, respectively, depending on the number of trace estimators used.

\subsubsection{Gradient estimation}
\textbf{Adjoint ODE.}
To compute the exact gradient $\nabla_{\mathbf{x}}\log p_\theta(\mathbf{x})$, we would need to backpropagate through the ODE solve. This is too memory-intensive, so we opt for the continuous adjoint method~\cite{chen2018neural,kidger2022neural}, which solves a secondary ODE that gives the gradient of the idealized continuous-time primary ODE. This adjoint method best balances our memory, speed, and accuracy requirements. Direct backpropagation through the probability flow ODE could be possible with improved gradient-checkpointing.

\subsection{Posterior Sampling Experiments}
\textbf{Gaussian ground-truth distribution.} The Gaussian distribution is defined for $16\times 16$ grayscale images. The mean and covariance were fit by expectation-maximization to images from the CelebA training set (each image was first center-cropped to $140\times 140$ and then rescaled to $16\times 16$). The covariance was preconditined by adding $0.01$ along the diagonal. To generate a batch of training data for a score model, samples are randomly drawn from the resulting Gaussian distribution.

\textbf{Score model.} All score models that were trained on $32\times 32$ images had an NCSN++ architecture~\cite{song2021scorebased} with 64 filters in the initial layer. The score model trained on the Gaussian ground-truth distribution in Main Sec.~3.2 had 128 filters in the initial layer.

\textbf{DPI implementation.}
We adapted the PyTorch implementation of DPI~\cite{sun2021deep}~\footnote{\url{https://github.com/HeSunPU/DPI}} for JAX/Flax. For all presented results on image posterior sampling, we used a RealNVP architecture with 64 affine-coupling layers. 
The RealNVP was optimized with stochastic gradient descent (SGD) with a batch size of 64. We used Adam optimizer with a learning rate of 0.0002 and clipped gradients to have norm 1.

\textbf{DPI sampling.} Once optimized, the RealNVP can be sampled to obtain samples from the approximate posterior. Occasionally the RealNVP produces a clearly out-of-distribution sample, so we remove such outliers by discarding any sample with a pixel value whose magnitude is greater than $2$. Although not needed in most cases, we applied this postprocessing step before computing statistics of DPI-estimated posteriors. 

\textbf{DPI optimization time.}
The main computational bottleneck is computing log-probabilities for each batch. Since we use adaptive step-size controllers, the time required for each SGD step is variable. In our experiments, we found it ranged from 30 seconds/step to 200 seconds/step. The time required for each ODE solve could also depend on the complexity of the distribution underlying the score-based prior. For example, we found CelebA priors to be faster (about 50 seconds/step for interferometric imaging experiments) and CIFAR-10 priors to be slower (about 200 seconds/step for deblurring a CIFAR-10 image, which was the slowest case). The RealNVP generally converges within 5000-10000 SGD steps, although we ran the optimization for 20000-50000 steps to be sure of convergence. We used v4-8 TPUs to perform the optimization.

Although DPI with a score-based prior takes a long time to optimize, it is extremely efficient for \textit{sampling}. Sampling 128 samples ($32\times 32$ RGB images) takes about 2.76 seconds. In contrast, the diffusion-based baselines that we include in the main text are much slower. To get 128 samples, SDE+Proj~\cite{song2022solving} takes 20.8 seconds; Score-ALD~\cite{jalal2021robust} (with 5 Langevin-dynamics steps at each annealing level) takes 51.8 seconds; and DPS~\cite{chung2023diffusion} takes 34.1 seconds.

Furthermore, our framework gives a reliable and rich posterior automatically. This saves human time and effort that would have been spent on carefully handcrafting and validating regularizers/priors.

\subsection{2D Experiments}
Supp.~Fig.~\ref{fig:2d_example} and Main Fig.~4 compare our posterior-sampling approach to baselines (SDE+Proj, Score-ALD, DPS) on a toy 2D posterior. Our samples were generated from a RealNVP with 32 affine-coupling layers. All methods used the same true score model. Since baseline methods do not provide posterior probabilities (only samples), we used kernel density estimation (KDE) to approximate a probability density function (PDF) from 10000 samples. In Fig.~\ref{fig:2d_example}, PDFs were estimated with \texttt{scipy.stats.gaussian\_kde}, which includes automatic KDE bandwidth selection. In Main Fig.~4, \texttt{sklearn.neighbors.KernelDensity} was used with a bandwidth-0.03 Gaussian kernel, since \texttt{scipy.stats.gaussian\_kde} does not do well on multimodal distributions.

\section{Image-Restoration Metrics}
\begin{figure}[!ht]
    \centering
    \includegraphics[width=0.48\textwidth]{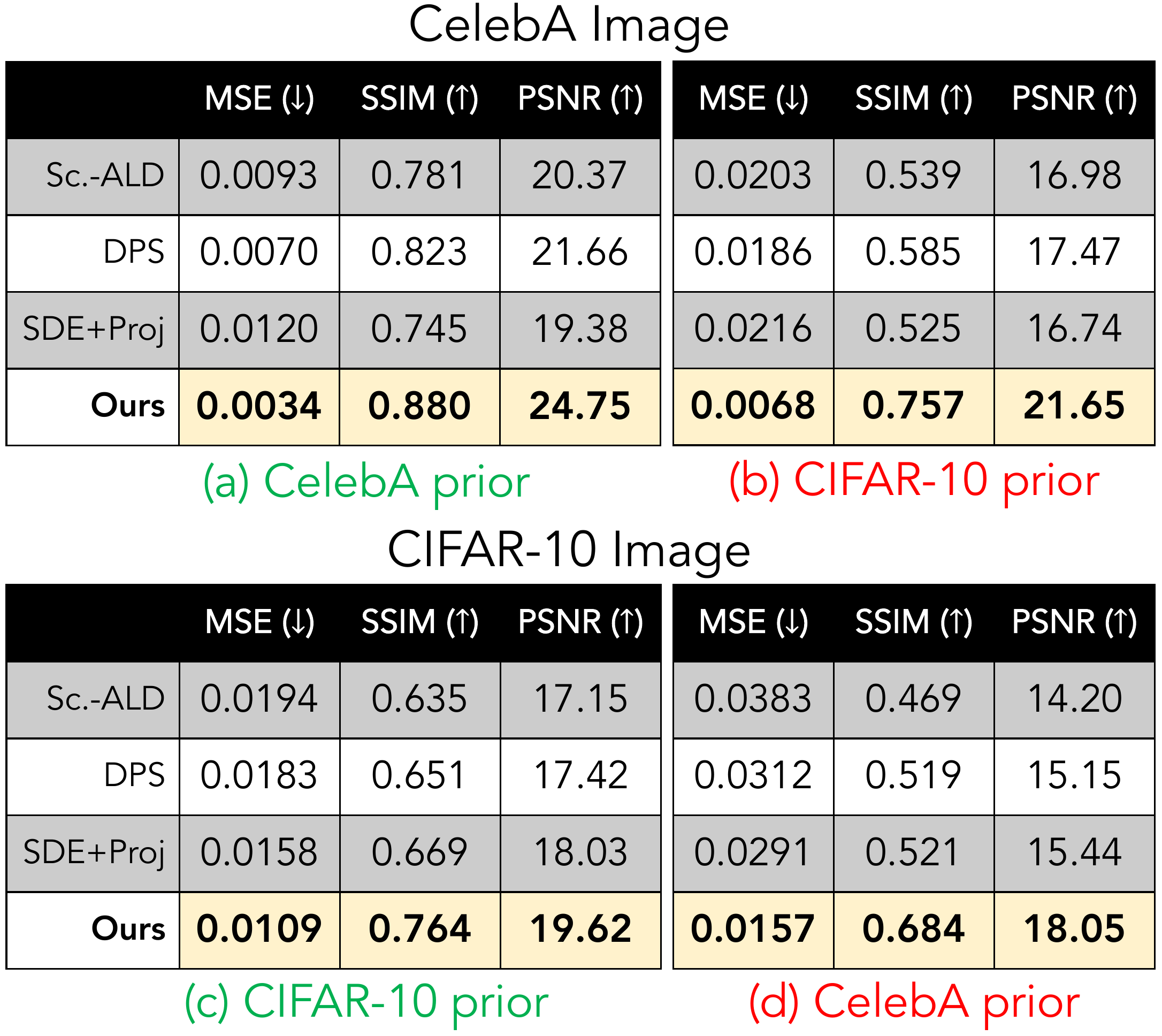}
    \caption{Image-restoration metrics (deblurring example). For Main Figs.~7 and 8, average MSE, SSIM, and PSNR of 128 estimated samples from each method compared to the true source image were evaluated. Our posterior samples outperform baseline samples for every combination of a source image and prior (e.g., CIFAR-10 prior applied to a CelebA source image).}
    \label{fig:deblurring_metrics}
\end{figure}

For our results on deblurring, we evaluated our chosen posterior-sampling approach against baselines (SDE+Proj, Score-ALD, DPS) using standard image-restoration metrics (MSE, SSIM, PSNR). Fig.~\ref{fig:deblurring_metrics} shows the evaluated metrics. We emphasize that such metrics do not reflect the correctness of the posterior. But for applications that call for high-quality posterior samples, Fig.~\ref{fig:deblurring_metrics} suggests that our framework is still preferable to baselines. 

\section{Score-Based Priors vs. Discrete-Flow Priors}
Although discrete normalizing flows (e.g., RealNVP~\cite{dinh2016density}, Glow~\cite{kingma2018glow}) are generative networks that provide image probabilities, they suffer from two limitations: (1) they are restricted to invertible network architectures, which limits the ability to express a diverse and sophisticated image distribution; and (2) their probability function does not generalize well outside of training data. We note that a score-based diffusion model (following the probability flow ODE) is actually a \textit{continuous}-time normalizing flow~\cite{song2021scorebased}.

\begin{figure}[!ht]
    \centering
    \includegraphics[width=0.36\textwidth]{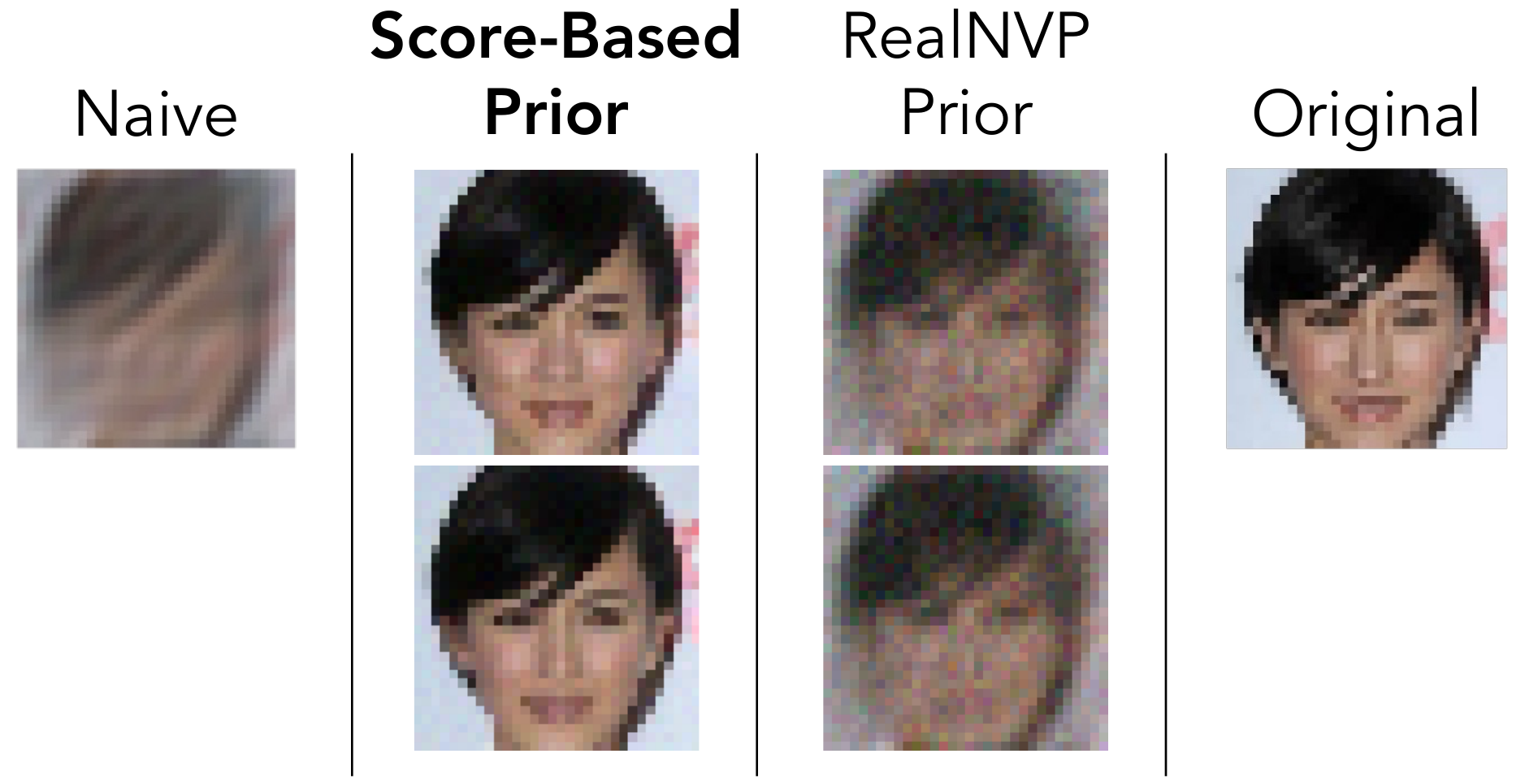}
    \caption{Score-based prior vs.~RealNVP prior. A score-based diffusion model and a RealNVP were each trained on the same CelebA training set. We then applied each of their probability functions as the prior in DPI for the task of deblurring (the same task as in Main Fig.~7). Two samples are shown from each estimated posterior.}
    \label{fig:realnvp_deblurring}
\end{figure}

\begin{figure}[tb]
    \centering
    \includegraphics[width=0.5\textwidth]{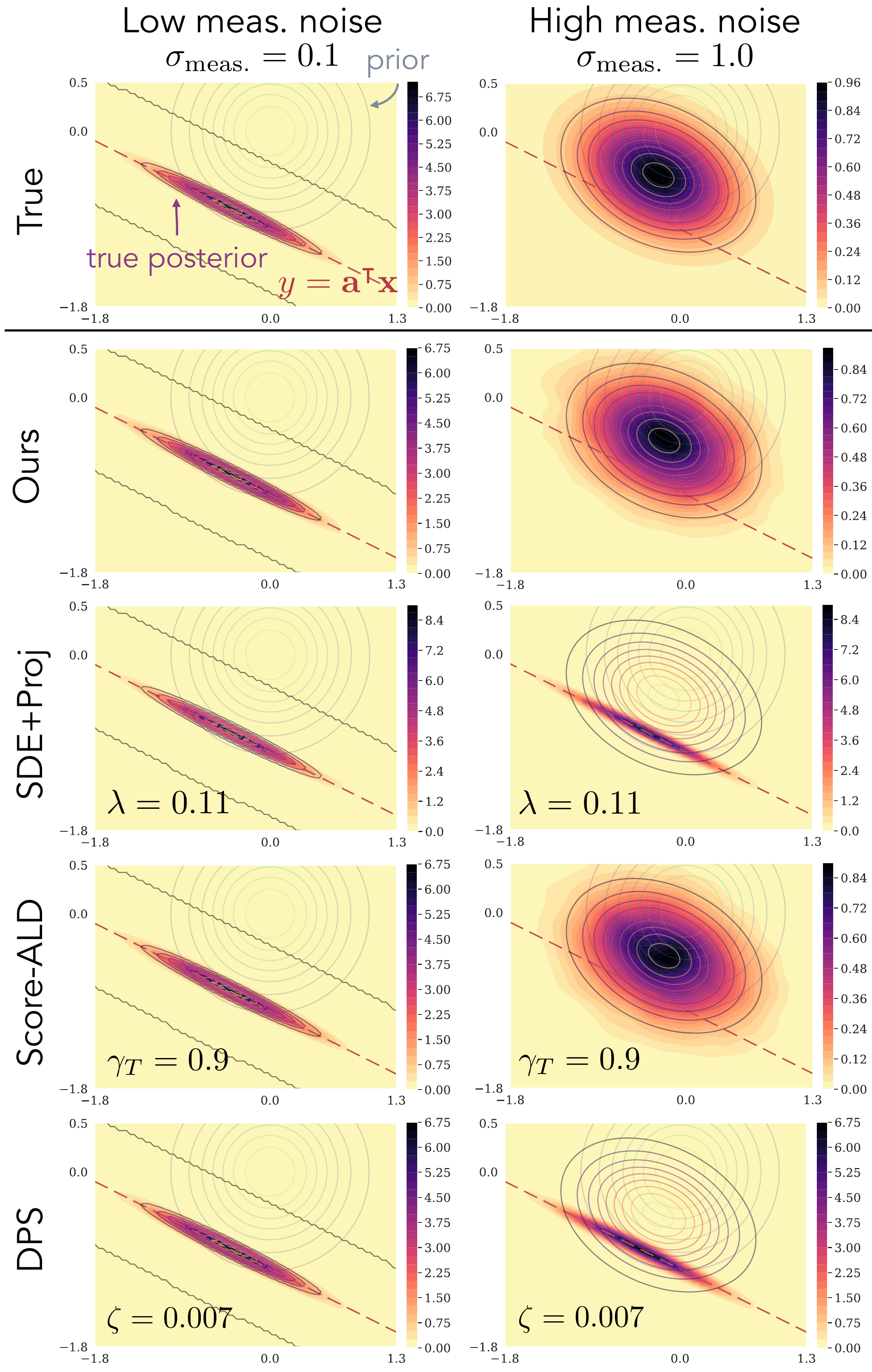}
    \caption{In this toy example with a Gaussian prior and linear measurements, we see that our method samples from the true posterior regardless of the measurement noise. For SDE+Proj, Score-ALD, and DPS, the optimal hyperparameter value was found (according to sample-approximated KL divergence to the true posterior) for the ``Low meas.~noise'' case. The same value was then applied for the ``High meas.~noise'' case. SDE+Proj and DPS severely under-estimate the spread of the posterior. Score-ALD works well here since the approximation of the measurement-likelihood converges to the true likelihood distribution as ALD continues. In general, though, Score-ALD can become unstable when the measurement annealing rate (i.e., the sequence of $\gamma_t$'s) is not well-tuned.}
    \label{fig:2d_example}
\end{figure}

Main Fig.~6 and Main Tab.~1 show the results of using a discrete normalizing-flow (NF) image prior for denoising. Fig.~\ref{fig:realnvp_deblurring} shows results for deblurring.
In both experiments, the NF used was a RealNVP with 64 affine-coupling layers, and DPI optimization was done with a learning-rate of $10^{-5}$ and gradients clipped to a norm of $1$. Compared to the score-based prior trained on the same dataset, the NF prior resulted in less visually-convincing samples and caused unstable optimization of the DPI posterior. The inferior quality of samples might be due to the limited expressiveness of a discrete NF. The instability might be due to the NF's inability to generalize to non-training images. This is relevant for inference algorithms that are randomly initialized (like DPI), as randomly-initialized images are likely far away from the prior. We note that, although an NF prior consistently performed poorly in our experiments, clever initialization might make DPI optimization more stable with an NF prior, and other NF architectures might be more expressive than a RealNVP architecture.

While a RealNVP is not as expressive as a diffusion model, our experiments with DPI suggest that a RealNVP can model a \textit{posterior} that is sufficiently constrained by measurements. If the inverse problem is extremely ill-posed --- meaning the posterior is almost indistinguishable from the prior --- then a RealNVP would probably not sufficiently capture the distribution. DPI is not restricted to discrete normalizing flows, though. As long as the generative model used to approximate the posterior is invertible, it can be optimized via the variational objective.

{\small
\bibliographystyle{ieee_fullname}
\bibliography{supp}
}